# 3LM: Bridging Arabic, STEM, and Code through Benchmarking

**Basma El Amel Boussaha, Leen AlQadi, Mugariya Farooq, Shaikha Alsuwaidi**
**Giulia Campesan, Ahmed Alzubaidi, Mohammed Alyafeai, Hakim Hacid**
Technology Innovation Institute, Abu Dhabi, UAE
basma.boussaha@tii.ae

## Abstract

Arabic is one of the most widely spoken languages in the world, yet efforts to develop and evaluate Large Language Models (LLMs) for Arabic remain relatively limited. Most existing Arabic benchmarks focus on linguistic, cultural, or religious content, leaving a significant gap in domains like STEM and code which are increasingly relevant for real-world LLM applications. To help bridge this gap, we present **3LM**, a suite of **three** benchmarks designed specifically for Arabic. The first is a set of STEM-related question-answer pairs, naturally sourced from Arabic textbooks and educational worksheets. The second consists of synthetically generated STEM questions, created using the same sources. The third benchmark focuses on code generation, built through a careful translation of two widely used code benchmarks, incorporating a human-in-the-loop process with several rounds of review to ensure high-quality and faithful translations. We release all three benchmarks publicly to support the growth of Arabic LLM research in these essential but underrepresented areas[1].

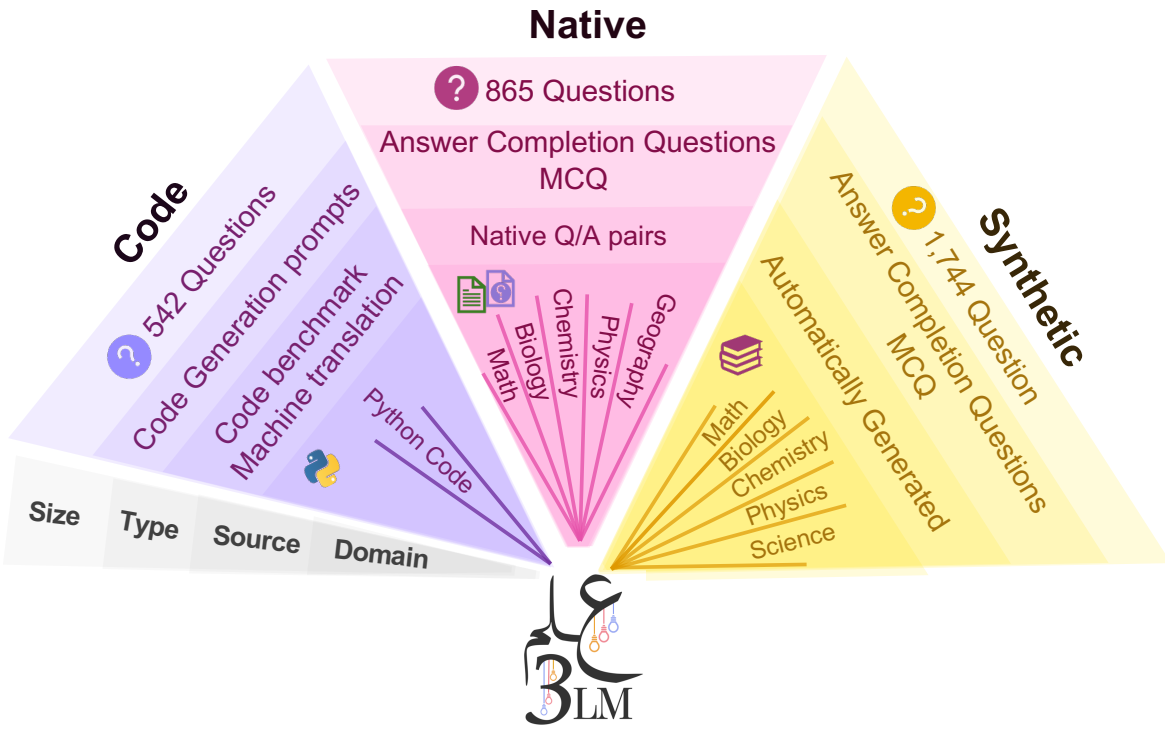

Figure 1: Summary of 3LM Benchmark.

## 1 Introduction

The rapid advancement of Large Language Models (LLMs) has underscored the critical need for high-quality, domain-specific evaluation benchmarks. While several benchmarks have recently been proposed for Arabic, many focus on specific linguistic or cultural dimensions such as dialectal variation (Mousi et al., 2025), religious and cultural contexts (Alwajih et al., 2025), or general Arabic language understanding (Almazrouei et al., 2023) or are translated adaptations of English benchmarks, such as ArabicMMLU (Sengupta et al., 2023).

Despite these efforts, there remains a notable gap in native, scientifically grounded benchmarks designed to evaluate Arabic LLMs in structured, knowledge-intensive domains like science and mathematics. To address this, we introduce **3LM** (علم), a suite of three benchmarks for evaluating Arabic LLMs across core STEM disciplines, including general science, mathematics, chemistry, physics, and biology, and code generation.

The first benchmark in 3LM consists of native multiple-choice questions (MCQs) sourced from real Arabic-language educational worksheets, textbooks, and other pedagogical content collected from various countries and regions. The second benchmark is synthetic, generated using the YourBench framework (Shashidhar et al., 2025) by HuggingFace, based on scientific textbooks and course materials crawled from Arabic educational platforms. The third benchmark adapts two established code and reasoning benchmarks MBPP and HumanEval via a rigorous machine translation pipeline that incorporates human-in-the-loop validation through multiple verification and correction stages.

The contributions of this paper are threefold: First, we present three comprehensive benchmarks spanning STEM domains and code generation, constructed through rigorous methodologies that ensure authenticity and quality from native Arabic content curation to synthetic generation and careful translation with human verification. Second, we conduct an extensive evaluation of over 40 state-of-the-art Arabic and multilingual LLMs, providing the most comprehensive assessment of Arabic language model capabilities in scientific and programming domains to date. Third, we perform thorough analysis including cross-task correlations and robustness testing, revealing insights into model behavior and the relationship between different cognitive capabilities in Arabic LLMs.

[1] 3LM benchmark is accessible on https://github.com/tiiuae/3LM-benchmark

By focusing on high-quality, natively Arabic, and scientifically relevant content, 3LM fills a key gap in the ecosystem of Arabic LLM evaluation, offering a more representative and robust framework for assessing model capabilities in formal knowledge domains.

## 2 Related Work

The development of Arabic language model evaluation has witnessed remarkable growth, with numerous initiatives addressing the unique challenges of assessing Arabic LLMs across diverse domains. AlGhafa (Almazrouei et al., 2023) pioneered a comprehensive evaluation by introducing a new MCQ benchmark for Arabic LLMs that evaluates models on a range of abilities, including reading comprehension, sentiment analysis, and question answering. ORCA (Elmadany et al., 2023) complemented these efforts by offering a comprehensive comparison between 18 multilingual and Arabic language models with a unified single-number evaluation metric.

Cultural understanding has been extensively explored through specialized benchmarks. Jawaher (Magdy et al., 2025) assessed cultural knowledge through Arabic proverbs, designed to assess LLMs' capacity to comprehend and interpret Arabic proverbs, including proverbs from various Arabic dialects. ArabicSense (Lamsiyah et al., 2025) focused on commonsense reasoning by testing whether systems can distinguish between natural language statements that make sense and those that do not. Additional cultural benchmarks include Arabic Culture (Sadallah et al., 2025), Palm (Alwajih et al., 2025), and Fann or Flop (Alghallabi et al., 2025), which captures multi-genre and multi-era variations.

Linguistic diversity has been addressed through Aradice (Mousi et al., 2025), focusing on dialectal variations, while specialized domains are covered by ArabLegalEval (Hijazi et al., 2024) for legal text understanding. Arabic MMLU (Nacar et al., 2025) attempted to adapt English benchmarks, although critical analysis revealed significant deficiencies, encompassing linguistic inconsistencies, semantic imprecisions, and fundamental methodological flaws.

Despite these valuable contributions, a critical gap exists in STEM evaluation. To the best of our knowledge, AraSTEM (Mustapha et al., 2024) represents the only dedicated STEM benchmark, introducing a new Arabic multiple-choice question dataset for evaluating LLMs knowledge in STEM subjects across different levels. However, this benchmark remains inaccessible despite promises of open-source release, creating a substantial limitation in evaluating Arabic language models' scientific capabilities.

On the other hand, code generation evaluation has been dominated by English-based benchmarks, with HumanEval (Chen et al., 2021) and MBPP (Austin et al., 2021) serving as gold standards. HumanEval comprises 164 human-generated tasks with function signatures, docstrings, and test cases, while MBPP contains 974 crowd-sourced Python programs with basic problem statements. Recent advances through EvalPlus (Liu et al., 2023) have addressed test coverage limitations, with HumanEval+ expanding test suites by 80× and MBPP+ providing 35× more tests, demonstrating superior capabilities in detecting incorrect code.

The growing importance of multilingual code evaluation stems from bilingual and multilingual models like JAIS (Sengupta et al., 2023) and AceGPT (Huang et al., 2024), which are trained on Arabic, English, and code content. Initial multilingual efforts include HumanEval-XL (Peng et al., 2024) and mHumanEval (Raihan et al., 2025), which extended HumanEval to multiple languages, including Arabic. However, these efforts focus solely on base benchmarks without enhanced test coverage and lack comprehensive treatment of MBPP, with MBXP (Athiwaratkun et al., 2023) addressing only programming language diversity while maintaining English prompts.

This landscape reveals that while existing benchmarks excel in cultural knowledge and general language understanding, there are urgent needs for comprehensive, open-source STEM and multilingual code evaluation tools. To address these critical gaps, we introduce **3LM**, a comprehensive benchmark suite comprising three novel Arabic evaluation datasets covering mathematics, physics, chemistry, biology, general science, and programming.

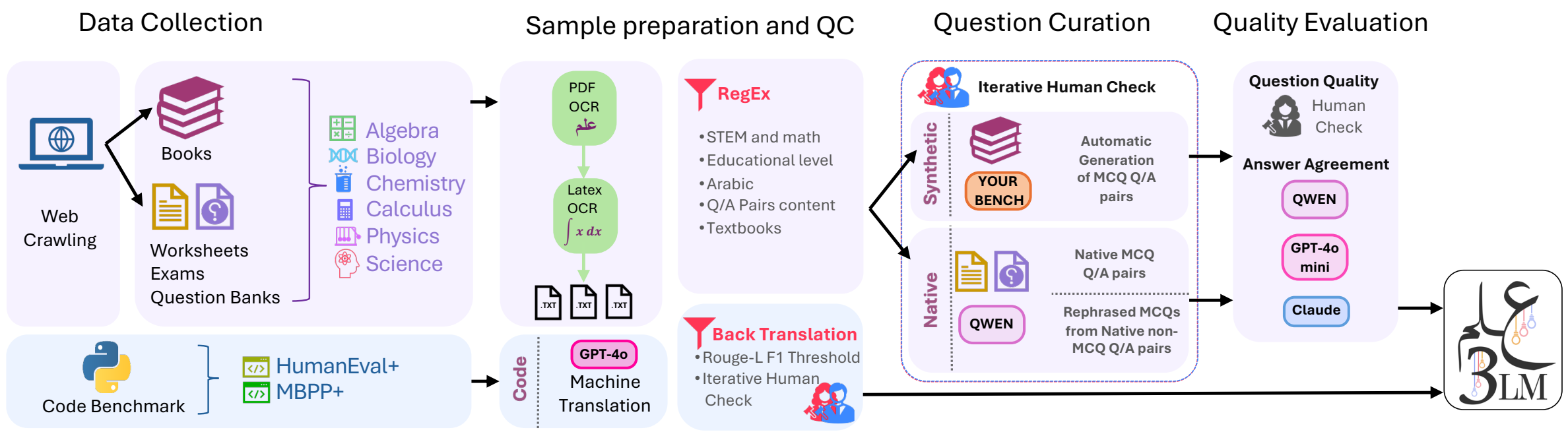

Figure 2: 3LM benchmark curation process.

Unlike previous efforts, 3LM is fully open-source with all datasets publicly available[2], accompanied by a comprehensive GitHub repository containing all the code necessary to reproduce the experimental results reported in this paper.

## 3 The Benchmark

3LM benchmark comprises two categories: STEM and code. The STEM portion includes both automatically generated synthetic questions from textbooks and native questions from various sources. Figure 1 illustrates a summary of the benchmarks, and Figure 2 outlines the key curation steps detailed in the following sections.

### 3.1 STEM

The construction process of the STEM benchmarks are detailed in the following subsections.

#### 3.1.1 Data Collection

Educational content was systematically collected from various online sources, including educational websites and open question banks, using web scraping, API calls, and targeted keyword searches. Only PDFs containing biology, chemistry, physics, general science, and mathematics content were retained. These PDFs were categorized using regex pattern matching based on the documents' titles.

Higher priority was given to PDFs with explicitly stated academic levels targeting *middle* and *high school* students, which filtered out image-heavy content designed for primary level students. The collected material focused on worksheets, exams, and question banks containing question-answer pairs suitable for OCR processing.

Given the prevalence of mathematical equations and complex notation in STEM content, a specialized Math-based OCR pipeline was employed. Pix2Tex (Blecher, 2023), a LaTeX OCR model, was used to accurately convert mathematical notation into LaTeX code. This dual-stage OCR process (see 2) resulted in a curated collection of over 1,081 pages of STEM content with structured question-answer pairs.

#### 3.1.2 Native benchmark

MCQs, spanning varying difficulty levels and covering authentic educational content, were extracted from text documents as described in Section 3.1.1.

The native benchmark construction follows a systematic four-stage pipeline using Qwen3-235B-A22B[3] to ensure high-quality contextually complete MCQ pairs:

**Question-Answer Extraction.** Each document was processed separately, with the model extracting complete question-answer pairs along with any necessary context. General instructions were added at the beginning when they applied to multiple questions, and when the answers were not explicitly labeled in the questions, they were extracted from an answer key.

**Classification and Filtering.** Extracted pairs underwent systematic classification across four dimensions: (1) *Question Type* (MCQ, Completion, Generative, Other); (2) *Difficulty Level* (1-10 scale); (3) *Domain Classification* (STEM subject areas); and (4) *Visual Dependency*. Questions requiring visual elements were eliminated since the OCR pipeline focused exclusively on textual content.

**Format Standardization.** The final stage achieved format consistency through: (1) removal of extraneous labels and formatting inconsistencies, and (2) conversion of non-MCQ questions into MCQ format. New MCQ versions included four options labeled (أ, ب, ج, د) with correct answers randomly assigned to avoid positional bias.

[2]Code: https://huggingface.co/datasets/tiiuae/evalplus-arabic
Synthetic: https://huggingface.co/datasets/tiiuae/SyntheticQA
Native: https://huggingface.co/datasets/tiiuae/NativeQA

[3]https://huggingface.co/Qwen/Qwen3-235B-A22B

**Quality Assurance.** All question-answer pairs underwent manual verification by the research team to ensure accuracy, coherence, and adherence to MCQ format requirements, and to validate the educational integrity and linguistic quality of the automated process..

Complete prompts for each stage of the pipeline are provided in Appendix B.

#### 3.1.3 Synthetic benchmark

Text sources from Section 3.1.1 were processed through a QA generation pipeline to synthetically generate domain-specific multiple-choice question-answer pairs. The YourBench (Shashidhar et al., 2025) pipeline was employed with modifications for Arabic content, including Arabic letters (أ, ب, ج, د) for answer choices instead of A,B,C,D.

The pipeline consists of five LLM-powered stages adapted for Arabic content:

**Ingestion.** Input documents are preprocessed and converted into structured Markdown format.

**Summarization.** Documents are summarized while removing metadata, redundant content, HTML tags, and web artifacts. The LLM identifies main topics and salient points while maintaining logical consistency and global context.

**Chunking.** Summarized text is segmented into semantically coherent chunks, creating both single-hop and multi-hop chunks for different reasoning levels.

**Question Generation.** Multi-hop chunks generate challenging multiple-choice questions requiring information synthesis across document parts. The LLM creates questions with four answer choices and assigns difficulty levels (1-10 scale). An embedding-based similarity mechanism identifies and manages closely related questions.

**Analysis.** QA pairs are evaluated for content coverage and question diversity.

From collected STEM books, multiple-choice QA pairs were synthetically generated across mathematics, physics, chemistry, biology, and general science. Seeded random sampling selected document chunks for question generation. Rigorous filtering removed QA pairs referencing visual artifacts, enforced a difficulty threshold of 6 or higher, and ensured high topical and structural diversity among final QA pairs.

### 3.2 Code

To assess the programming capabilities of bilingual and multilingual LLMs, we extend the EvalPlus leaderboard benchmarks to Arabic through refined machine translation.

Our approach translates HumanEval+[4] and MBPP+[5] datasets using GPT-4o. For HumanEval, only docstring descriptions are translated, preserving variables and test cases. For MBPP, the full prompt is translated as it consists of plain natural language task descriptions.

Translation quality is validated through rigorous backtranslation using the same GPT-4o model. ROUGE-L F1 scores between original English prompts and backtranslated versions establish quality thresholds of 0.85 for HumanEval and 0.8 for MBPP (distributions in Appendix A.5). Translations below these thresholds undergo human review by native Arabic speakers with Python programming expertise, ensuring both linguistic accuracy and technical precision.

This process yields HumanEval-Arabic (HumanEval-Ar) and MBPP-Arabic (MBPP-Ar) benchmarks in base and plus versions, constituting the EvalPlus-Arabic (EvalPlus-Ar) suite. System and response prompts are adapted (Appendix A.2) to maintain Arabic linguistic conventions while preserving technical requirements. Example prompts are provided in Appendix A.1.

## 4 Benchmarks Characteristics

In comparison to other Arabic benchmarks, 3LM targets STEM content with source material originally in Arabic.

**Benchmark Size.** After quality iterations, the benchmark comprises 865 *native* question-answer pairs, 1,744 automatically generated *synthetic* questions, and 542 high-quality machine-translated *code* prompts (Figure 1).

**Domain Distribution.** The native benchmark spans biology, chemistry, physics, math, and geography, while the synthetic benchmark covers biology, chemistry, physics, math, and general science (Figure 3c). The synthetic benchmark includes diverse question types (conceptual, analytical, factual,

[4] https://huggingface.co/datasets/evalplus/humanevalplus

[5] https://huggingface.co/datasets/evalplus/mbppplus

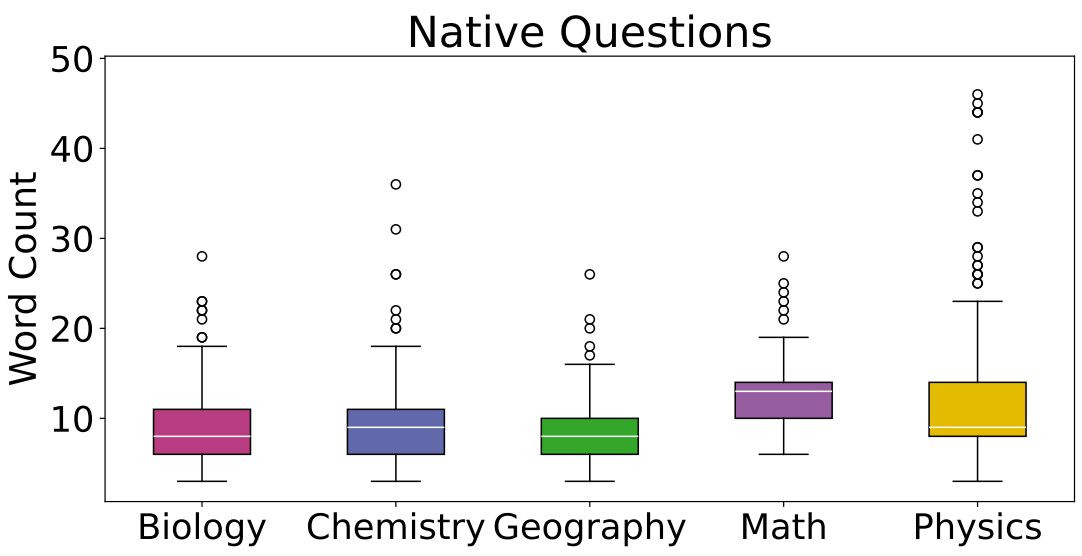

(a) Word count distribution in native benchmark.

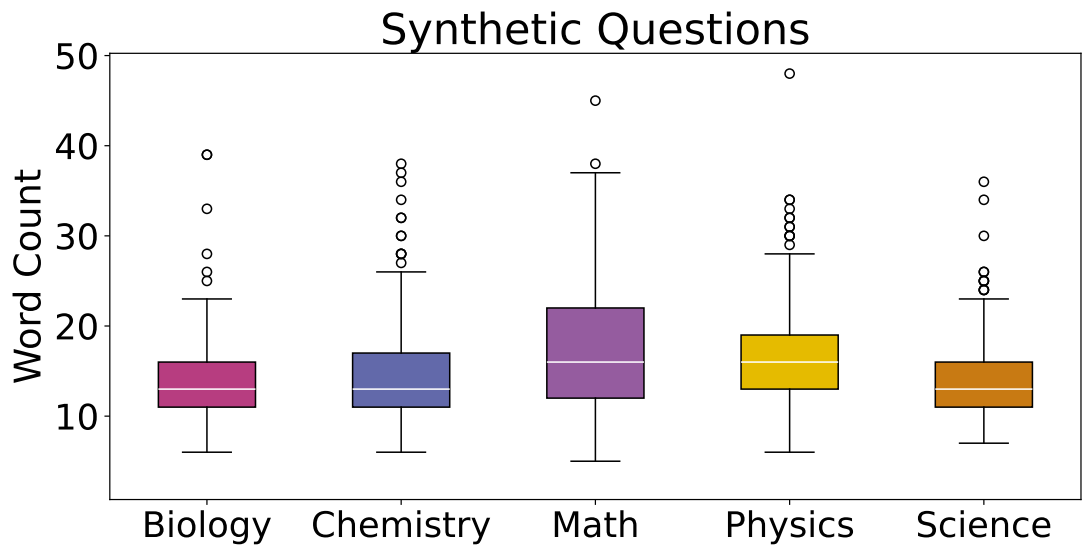

(b) Word count distribution in synthetic benchmark.

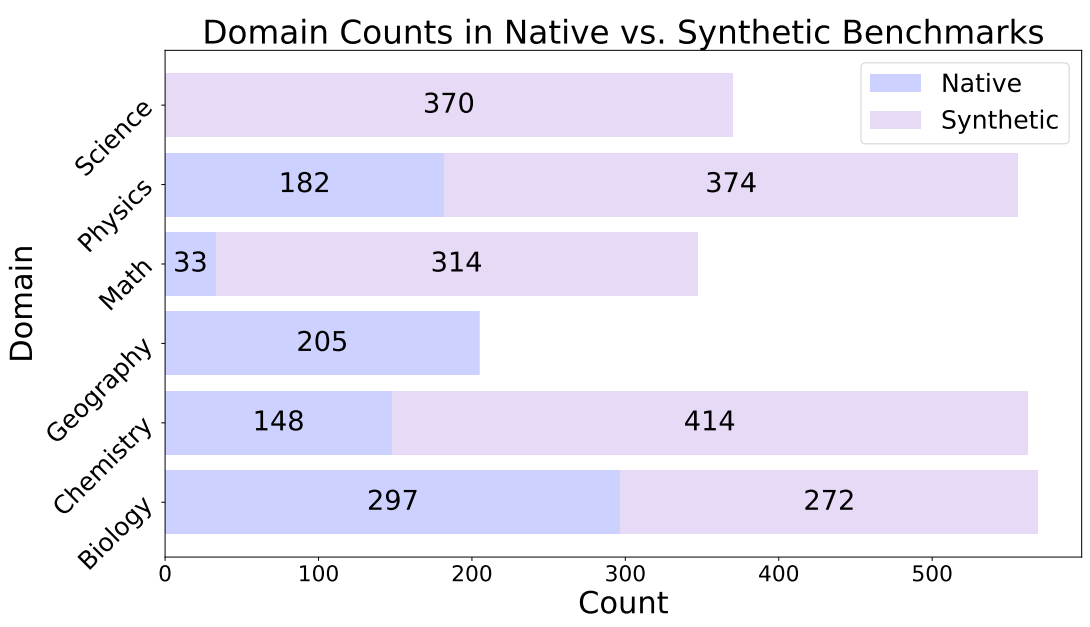

(c) Domain distribution in STEM benchmark.

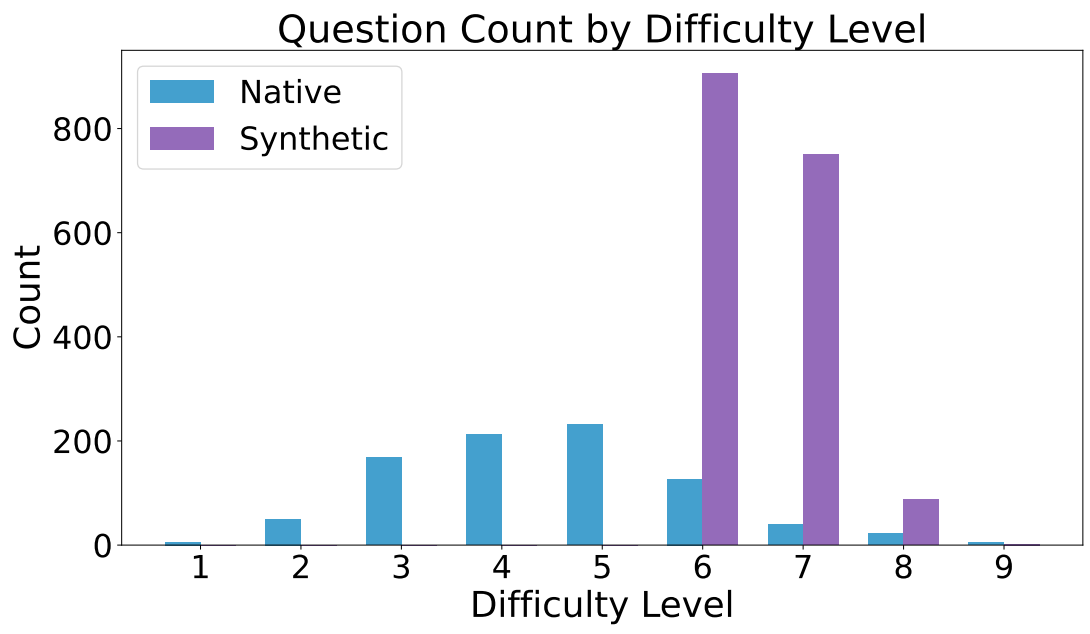

(d) Question difficulty distribution in STEM benchmarks.

Figure 3: Statistics on STEM benchmarks of 3LM.

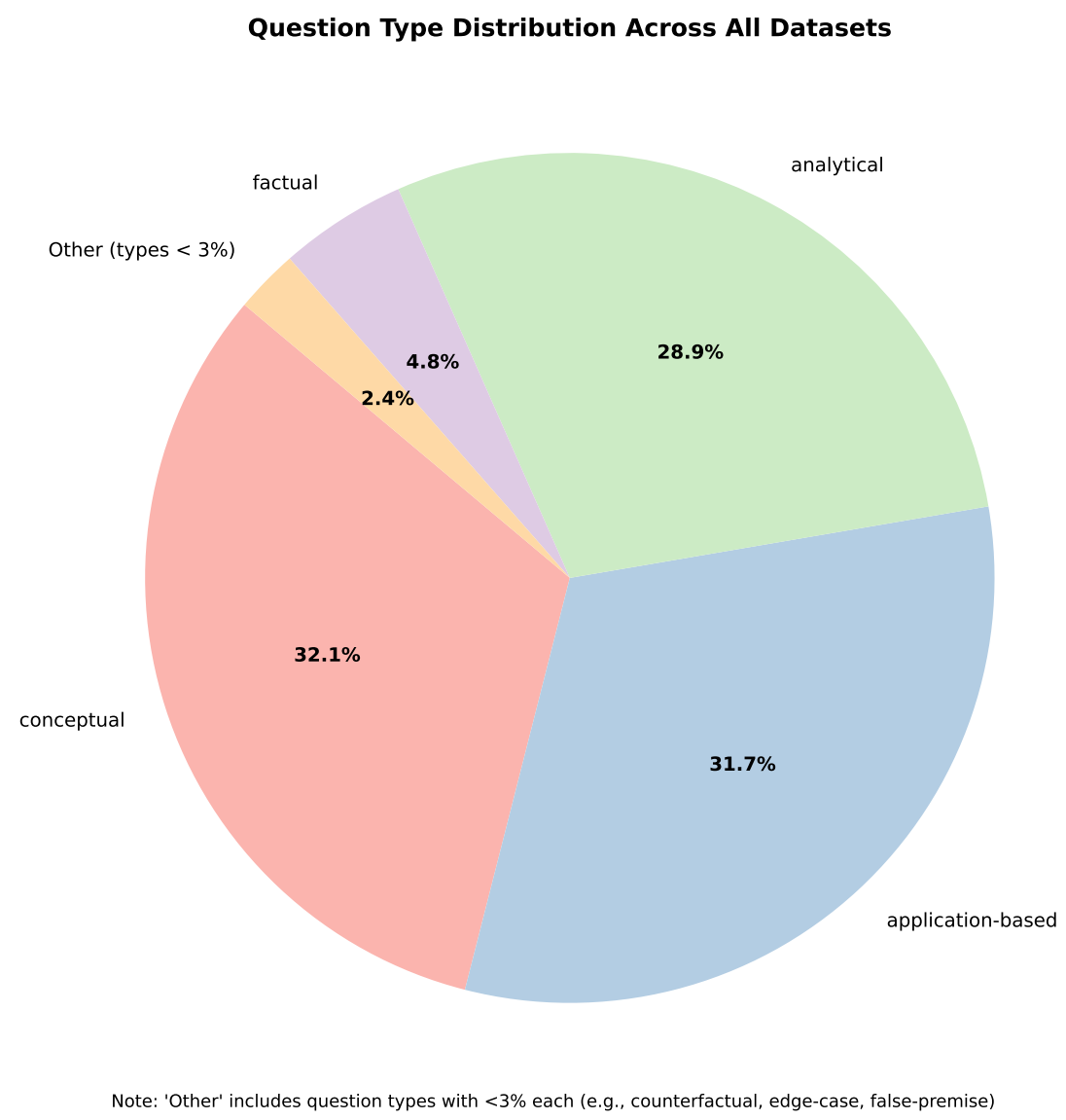

Figure 4: Question type distribution across domains in synthetic benchmark.

application-based) across domains. Figure 4 shows cross-dataset distributions, with per-domain question type distributions in Appendix C.

**Word Count Distribution.** Both native and synthetic prompts exhibit variety in word count, with maximum lengths of 48 words per question (Figure 3a and Figure 3b). Synthetic questions are generally longer, with math questions being the longest.

**Difficulty Distribution.** While source materials target middle and high school levels, LLM-estimated difficulty rankings show that native questions follow a Gaussian distribution with the challenge levels, whereas synthetic questions are consistently moderately to highly challenging (≥6) (Figure 3d).

**Code Benchmark Statistics.** The translated code benchmarks preserve EvalPlus scope while extending to Arabic. HumanEval-Arabic contains 164 prompts with 9.6 tests per task (base) and 748 tests per task (plus version, 80× expansion). MBPP-Arabic encompasses 378 prompts with 3 tests per task (base) and 105 tests per task (plus version, 35× expansion). Distribution plots are shown in Figure 8.

## 5 Experiments

In this section, we describe the experimental setup, the models, and the evaluation results.

### 5.1 Experimental Setup

We employ lighteval (Habib et al., 2023) for STEM benchmarks and evalplus (Liu et al., 2023) for code evaluation. Following Sadallah et al. (2025), STEM benchmarks were evaluated using two setups: (1) multiple-choice format, where models select from presented options, with accuracy computed based on Arabic letter likelihood (أ, ب, ج, د), and (2) completion format, where models generate answers to questions without visible choices, using joint likeli-hood of choice text with normalized accuracy for fairness across varying answer lengths. For code benchmarks, `pass@1` evaluation was adopted following the original HumanEval and MBPP benchmarks.

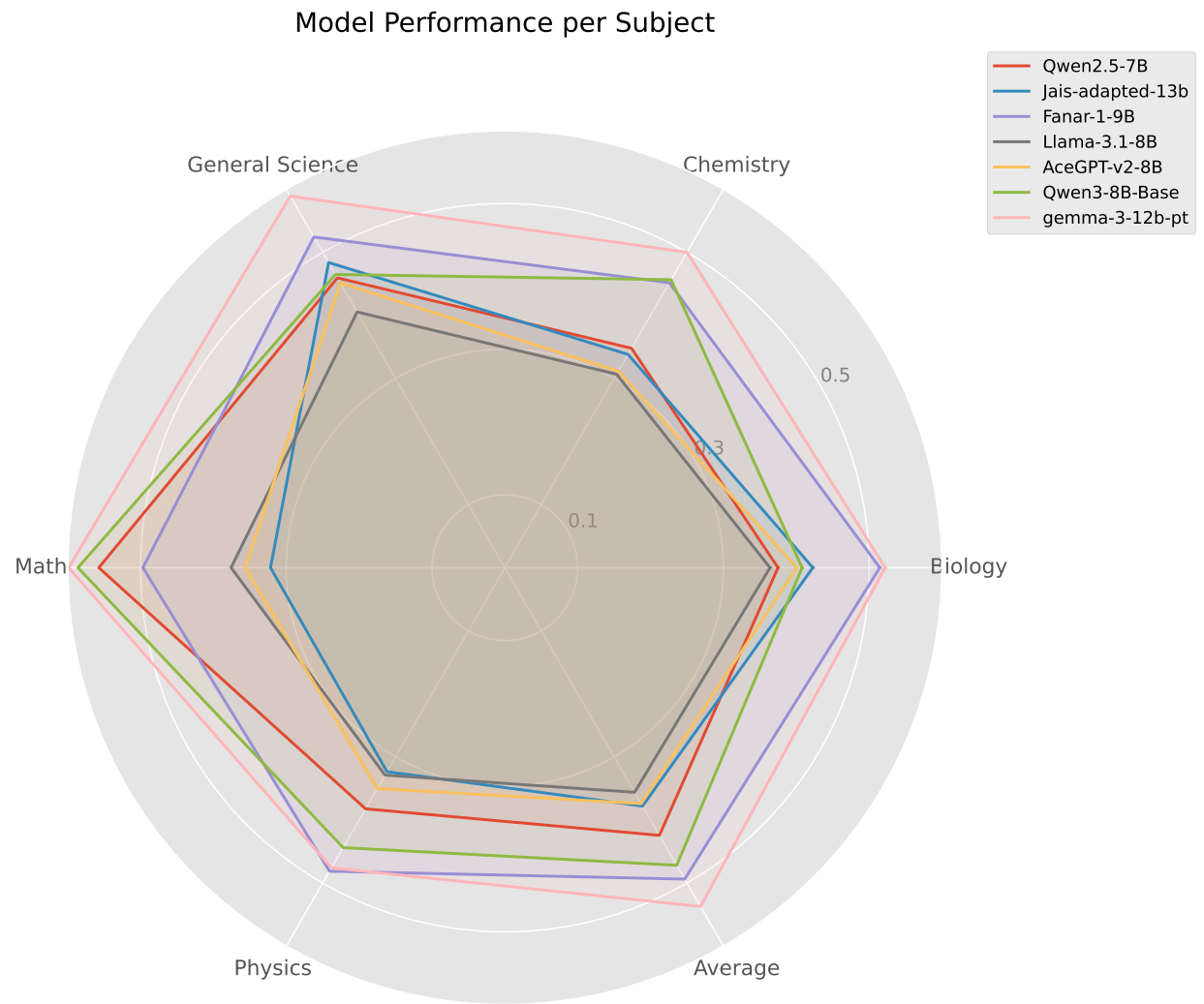

Figure 5: Subject-wise scores (completion) on base models ranging from 7B-13B.

| | | Native | | Synthetic | |
|---|---|---|---|---|---|
| Model | Size | MCQ | Completion | MCQ | Completion |
| Qwen2.5 | 7B | 86.13 | 48.43 | 79.5 | 42.19 |
| | 14B | 89.82 | 55.37 | 86.7 | 49.89 |
| | 32B | 93.41 | 56.18 | 89.9 | 50.09 |
| | 72B | **94.45** | <u>62.31</u> | **91.9** | 54.19 |
| jais-adapted | 13B | 43.81 | 57.91 | 40.6 | 40.14 |
| | 70B | 74.10 | 58.15 | 61.6 | 43.74 |
| jais-family-8k | 30B | 65.2 | 60.58 | 46.3 | 43.4 |
| Fanar-1 | 9B | 88.32 | 60.11 | 81.1 | 50.67 |
| Llama-3.1 | 8B | 73.52 | 45.78 | 63.8 | 35.44 |
| | 70B | 62.89 | 55.95 | 83.7 | 52.41 |
| AceGPT-v2 | 8B | 74.57 | 53.64 | 59.5 | 38.99 |
| | 32B | 81.27 | 55.95 | 68.9 | 40.24 |
| | 70B | 90.17 | 60.69 | 82.6 | 47.36 |
| Qwen3-Base | 4B | 87.05 | 48.32 | 78.6 | 41.13 |
| | 8B | 90.98 | 46.82 | 85.4 | 44.18 |
| | 14B | 87.98 | 50.98 | 84.0 | 50.37 |
| | 30B | <u>94.10</u> | 60.12 | <u>91.3</u> | <u>54.45</u> |
| gemma-3-pt | 4B | 81.15 | 52.02 | 68.4 | 40.78 |
| | 12B | 89.47 | 61.50 | 83.8 | 50 |
| | 27B | <u>94.10</u> | **67.63** | 89.8 | **59.42** |

Table 1: Average accuracy of MCQ vs. Completion for base models. **Bold** indicates the highest score in each column; <u>Underline</u> indicates the second best.

### 5.2 Models

Zero-shot evaluation was conducted across 40 models spanning various sizes, including both base and instruction-tuned variants. Multilingual model families include Gemma-3 (Team et al., 2025b), Llama-3 (Grattafiori et al., 2024), Qwen2.5 (Qwen et al., 2025), and Qwen3 (Yang et al., 2025). Arabic-centric families include AceGPT-v2 (Huang et al., 2024), Jais (Sengupta et al., 2023), and Fanar (Team et al., 2025a) for both transformers and Mixture of Experts (MoE) architectures[6].

### 5.3 Evaluation

In the following, the evaluation results of each of the Arabic LLMs on STEM and code benchmarks are provided.

#### 5.3.1 STEM

Models consistently perform better in MCQ format compared to completion format across all scales. Base model results for both evaluation formats are presented in Table 1, while instruction-tuned model results are reported in Table 8.

[6]Chat template enabled for instruct models.

For base models, as shown in Table 2 completion-based evaluation reveals counterintuitive performance patterns where larger models sometimes underperform compared to their smaller counterparts. `Gemma3-27B` dominates with top performance in 3 of 5 domains, while `Qwen3-30B-A3B` leads the remaining 2 domains. `Gemma3-27B` achieves the highest overall average across the benchmark. On the other hand, MCQ shows `Qwen2.5-72B` as the strongest performer, leading 3 of 5 domains. The MoE variant of Qwen excels in physics, while `Gemma3-27B` maintains its advantage in mathematics. Performance varies significantly by evaluation format and subject area.

For instruct models (Table 9, completion-based results show `Gemma3-27B` achieving the highest overall average, with `Qwen2.5-72B` as a close second. MCQ evaluation demonstrates `Qwen2.5-72B`'s consistent strength across all domains, with its 32B variant also performing competitively. Overall, models performance on native benchmark surpasses synthetic benchmark and this might be due to the difficulty level of the synthetic benchmark (Figure 3d. Figure 5 illustrates domain-wise performance for models in the 7B–13B parameter range under completion-based evaluation.

#### 5.3.2 Code

The same Arabic LLMs were evaluated on both the established EvalPlus (English) and novel EvalPlus-

| Model | Size | MCQ | | | | | Completion | | | | |
|---|---|---|---|---|---|---|---|---|---|---|---|
| | | Biology | Chemistry | General Science | Math | Physics | Biology | Chemistry | General Science | Math | Physics |
| Qwen2.5 | 7B | 84.9 | 72.2 | 85.1 | 77.4 | 77.8 | 37.13 | 35.51 | 49.19 | 50.64 | 38.5 |
| | 14B | 88.6 | 82.1 | 91.9 | 84.7 | 86.1 | 48.9 | 46.62 | 51.62 | 56.05 | 46.26 |
| | 32B | 93.4 | 87.4 | 92.9 | 85.0 | <u>90.6</u> | 50.37 | 46.14 | 51.35 | 56.05 | 46.52 |
| | 72B | <u>95.2</u> | <u>90.8</u> | **94.6** | **86.0** | **93.0** | 52.21 | 49.76 | 56.49 | <u>59.55</u> | 52.94 |
| jais-adapted | 13B | 43.0 | 38.2 | 46.5 | 34.7 | 40.4 | 43.75 | 36.23 | 51.08 | 34.08 | 35.56 |
| | 70B | 72.4 | 58.2 | 72.2 | 48.1 | 57.0 | 49.26 | 42.51 | 51.35 | 35.99 | 39.57 |
| jais-family-8k | 30B | 56.3 | 45.9 | 55.4 | 35.7 | 38.5 | 47.43 | 39.86 | 54.32 | 35.03 | 40.37 |
| QCRI/Fanar-1 | 9B | 89.0 | 80.4 | 87.6 | 69.4 | 79.1 | 53.31 | 47.1 | 55.14 | 48.09 | 49.73 |
| Llama-3.1 | 8B | 67.6 | 63.8 | 73.2 | 53.5 | 60.7 | 37.87 | 30.68 | 41.08 | 32.8 | 34.76 |
| | 70B | 92.3 | 81.9 | 90.3 | 72.0 | 82.1 | <u>55.51</u> | 52.66 | 54.86 | 50.64 | 48.4 |
| AceGPT-v2 | 8B | 65.8 | 60.9 | 69.7 | 45.2 | 55.9 | 43.38 | 33.09 | 44.32 | 35.67 | 38.5 |
| | 32B | 71.7 | 69.8 | 74.3 | 60.2 | 68.7 | 42.28 | 36.23 | 47.84 | 39.81 | 35.03 |
| | 70B | 90.4 | 81.6 | 90.8 | 69.7 | 80.5 | 50 | 45.41 | 52.97 | 46.18 | 42.25 |
| Qwen3 | 4B | 80.5 | 77.8 | 85.1 | 73.2 | 76.5 | 35.66 | 39.13 | 43.51 | 44.59 | 42.78 |
| | 8B | 85.7 | 84.8 | 91.4 | 81.2 | 84.0 | 41.18 | 42.51 | 46.49 | 47.13 | 43.58 |
| | 14B | 88.2 | 84.8 | 86.2 | 78.0 | 82.9 | 44.85 | 49.52 | 51.35 | 54.78 | 51.34 |
| | 30B-A3B | 94.1 | **92.5** | 93.8 | <u>85.7</u> | 90.4 | 50.37 | <u>54.35</u> | 52.43 | 59.24 | **55.88** |
| gemma-3-pt | 4B | 77.6 | 63.8 | 77.6 | 60.2 | 63.1 | 39.34 | 35.27 | 46.22 | 42.99 | 40.11 |
| | 12B | 91.9 | 81.4 | 90.3 | 73.2 | 82.1 | 51.84 | 51.69 | <u>57.57</u> | 53.5 | 51.87 |
| | 27B | **96.0** | 86.7 | <u>94.3</u> | 84.4 | 87.7 | **56.62** | **60.63** | **60.81** | **63.69** | <u>55.35</u> |

Table 2: Base models performance on the synthetic benchmark (values in percentages). **Bold** indicates the highest score in each column; <u>Underline</u> indicates the second highest.

Arabic suites were evaluated. All models use greedy generation with a maximum of 768 new tokens at 16-bit precision[7]. Instruct models include chat templates and system prompts, while reasoning models disable thinking mode. We report pass@1 scores (Chen et al., 2021). For base models, `Qwen3-14B-Base` achieves the highest average scores on both EvalPlus and EvalPlus-Ar benchmarks (Table 3). The top-5 positions are dominated by Qwen series models across both suites, reflecting their high-quality code training data (Qwen et al., 2025).

For instruct models, `Qwen3-30B-A30B` and `Qwen3-14B` deliver the best average performance despite not being the largest models evaluated (Table 10). Both Qwen and Gemma-3 series maintain competitive performance across their full size ranges. For the Arabic suite, Qwen2.5-72B-Instruct and Qwen3-32B achieve the highest scores.

The substantial performance gap between base and plus versions underscores the importance of comprehensive unit test coverage in code benchmarks. In addition to these evaluations, an in-depth study of the correlation between Arabic code generation, English code generation, and NLP tasks was conducted for a series of LLMs.The scores and the findings are reported in section A.4 in Appendix A.

[7] fp16 for JAIS series

# 6 Robustness under Distractor Perturbation

To evaluate models' reasoning capabilities and resistance to superficial pattern matching, 25% of Native Benchmark samples were systematically modified through targeted distractor manipulations. This *Robustness under Distractor Perturbation* (RDP) analysis tests three critical aspects: genuine STEM comprehension versus pattern matching, metacognitive awareness of insufficient information, and robustness to answer set variations.

**Methodology:** Two perturbation strategies were applied: (1) removed correct answers from 20% of samples, replacing them with Arabic phrases meaning "none of the above," and (2) introduced these phrases as additional distractors in 5% of samples by replacing incorrect choices. To prevent simple pattern matching, we randomly varied the Arabic expressions using semantically equivalent alternatives:

(1) لا شيء مما ذكر *(Nothing from what was mentioned)*
(2) ليس أيٌّ مما سبق صحيحًا *(None of the above is correct)*
(3) جميع ما سبق غير صحيح *(All of the above is incorrect)*
(4) لا شيء مما سبق *(Nothing from the above)*
(5) ليس أيٌّ مما ذكر صحيحًا *(None of what was mentioned is correct)*

This experimental design distinguishes between models that genuinely understand STEM concepts and those that rely on superficial matching strategies, while simultaneously assessing their ability to recognize when presented options lack correct answers which remains a crucial metacognitive skill

| Model | Size | English HumanEval | HumanEval+ | MBPP | MBPP+ | Arabic HumanEval | HumanEval+ | MBPP | MBPP+ | Average English | Arabic |
|---|---|---|---|---|---|---|---|---|---|---|---|
| Qwen2.5 | 7B | 58.5 | 50.0 | 77.2 | 64.3 | 50.6 | 42.7 | 70.6 | 57.7 | 62.5 | 55.4 |
| | 14B | 62.8 | 55.5 | 73.0 | 60.1 | 51.8 | 45.7 | 71.2 | 58.7 | 62.9 | 56.9 |
| | 32B | 57.9 | 52.4 | 83.3 | 69.0 | 65.7 | 51.8 | 47.0 | **82.8** | 67.2 | 62.2 |
| | 72B | 59.8 | 51.8 | **87.6** | **71.7** | <u>67.7</u> | <u>57.9</u> | 60.1 | 47.4 | 67.7 | 58.3 |
| jais-adapted | 13B | 18.9 | 13.4 | 31.5 | 24.6 | 13.4 | 9.8 | 29.1 | 22.8 | 22.1 | 18.8 |
| | 70B | 27.4 | 24.4 | 43.1 | 34.7 | 22.0 | 18.9 | 40.5 | 33.9 | 32.4 | 28.8 |
| jais-family-8k | 30B | 26.8 | 23.2 | 46.6 | 38.1 | 23.8 | 20.1 | 12.4 | 10.3 | 33.7 | 16.7 |
| QCRI/Fanar-1 | 9B | 32.9 | 29.3 | 64.3 | 51.9 | 31.7 | 25.6 | 60.8 | 49.5 | 44.6 | 41.9 |
| Llama-3.1 | 8B | 39.0 | 32.3 | 60.8 | 51.3 | 29.9 | 24.4 | 54.5 | 44.4 | 45.9 | 38.3 |
| | 70B | 56.7 | 50.0 | 78.3 | 66.7 | 49.4 | 40.9 | 70.4 | 59.8 | 62.9 | 55.1 |
| AceGPT-v2 | 8B | 33.5 | 28 | 57.9 | 47.1 | 28.1 | 23.8 | 50.8 | 40.7 | 41.6 | 35.9 |
| | 32B | 43.3 | 38.4 | 58.5 | 49.5 | 28.0 | 23.2 | 52.6 | 43.4 | 47.4 | 36.8 |
| | 70B | 47.0 | 38.4 | 64.8 | 55.6 | 42.1 | 36.0 | 54.5 | 45.2 | 51.5 | 44.5 |
| Qwen3-Base | 4B | 63.4 | 55.5 | 75.1 | 64.0 | 56.7 | 50.0 | 68.8 | 58.2 | 64.5 | 58.4 |
| | 8B | 69.5 | <u>63.4</u> | 76.2 | 64.0 | 63.4 | 56.7 | 74.6 | 61.9 | 68.3 | 64.2 |
| | 14B | **72.0** | **64.0** | <u>84.9</u> | <u>71.4</u> | **70.7** | **63.4** | **78.3** | <u>64.6</u> | **73.1** | **69.3** |
| | 30B-A3B | <u>70.7</u> | **64.0** | 84.7 | 68.5 | 65.2 | <u>57.9</u> | <u>78.0</u> | 63.5 | <u>72.0</u> | <u>66.2</u> |
| gemma-3-pt | 4B | 33.5 | 28.0 | 60.6 | 51.9 | 26.2 | 22.0 | 54.0 | 43.9 | 43.5 | 36.5 |
| | 12B | 47.0 | 38.4 | 73.8 | 61.1 | 35.4 | 29.3 | 66.7 | 54.8 | 55.1 | 46.6 |
| | 27B | 47.6 | 40.9 | 75.1 | 62.2 | 43.3 | 37.8 | 71.2 | 58.2 | 56.5 | 52.6 |

Table 3: Base models performance on the EvalPlus suite. **Bold** indicates the highest score in each column; <u>Underline</u> indicates the second best.

for real-world applications[8].

Experimental results on base models are given in Table 11 whereas instruct models are evaluated in Table 12. A consistent performance drop is observed under RDP perturbations, with base models showing larger accuracy declines than instruct-tuned ones. Notably, large instruct models (e.g. `Qwen2.5-72B` and `Llama-3.3-70B`) remain relatively stable, indicating stronger generalization and robustness to distractors. These trends emphasize the value of instruction tuning and highlight RDP as an effective probe for assessing authentic reasoning versus superficial pattern recognition.

## 7 Limitations

While 3LM provides comprehensive evaluation across STEM and coding domains, several limitations should be acknowledged. The benchmark primarily targets middle and high school-level content, potentially limiting assessment of advanced university-level scientific concepts and graduate-level research topics.

The synthetic benchmark generation process introduces potential biases inherited from the underlying language models such as `Qwen3-235B-A22B` used for question creation, which may reflect training data limitations or model-specific reasoning patterns. These biases could influence question difficulty, topic coverage, and answer distributions.

In the code benchmark, while natural language prompts are translated to Arabic, the variable names, and function signatures remain in English. This mixed-language approach may not fully capture the challenges faced by models when processing entirely Arabic-based programming contexts.

Finally, the benchmark is exclusively text-based, excluding visual elements such as diagrams, graphs, charts, and mathematical figures that are integral to many STEM domains. This limitation may underestimate the complexity of real-world scientific problem-solving that often requires visual reasoning and interpretation.

## 8 Conclusion

We introduce **3LM**, a comprehensive benchmark suite addressing the critical gap in Arabic STEM and code evaluation for large language models. Through systematic curation processes involving native content extraction, synthetic question generation, and machine translation with rigorous quality validation, we have created three complementary benchmarks spanning mathematics, physics, chemistry, biology, general science, and programming domains. Our extensive evaluation across multiple model architectures demonstrates the benchmark's effectiveness in revealing strengths and weaknesses in Arabic scientific reasoning and bilingual code generation capabilities.

To foster reproducible research and community engagement, we release 3LM as a fully open-source resource, complete with all datasets, evaluation code, and detailed documentation necessary to reproduce the experimental results presented in this work. We hope this contribution will encourage the Arabic NLP community to leverage these benchmarks for model development, comparative analysis, and future research directions, ultimately advancing the state of Arabic language models in scientific and technical domains.

[8]NativeQA-RDP: https://huggingface.co/datasets/tiiuae/NativeQA-RDP

# A Code Benchmark

## A.1 Example Prompts

**HumanEval\18**

**HumanEval**

```
def how_many_times(string: str,
substring: str) -> int:
    """ Find how many times a given
substring can be found in the original
string. Count overlapping cases.
    >>> how_many_times('', 'a')
    0
    >>> how_many_times('aaa', 'a')
    3
    >>> how_many_times('aaaa', 'aa')
    3
    """
```

**HumanEval-Ar**

```
def how_many_times(string: str,
substring: str) -> int:
    """أوجد عدد المرات التي يمكن أن يظهر فيها نص معين
داخل النص الأصلي. احسب الحالات المتداخلة.
    >>> how_many_times('', 'a')
    0
    >>> how_many_times('aaa', 'a')
    3
    >>> how_many_times('aaaa', 'aa')
    3
    """
```

**MBPP\18**

**MBPP**

```
Write a function to remove characters
from the first string which are present
in the second string.
```

**MBPP-Ar**

اكتب دالة لحذف الأحرف من السلسلة الأولى الموجودة في السلسلة الثانية.

## A.2 Instruction and Response Prompt

The instruction prompt is adapted from ”Please provide a self-contained Python script that solves the following problem in a markdown code block:” to ”يرجى تقديم برنامج بايثون مستقل يحل المشكلة التالية داخل كتلة تعليمات برمجية بتنسيق markdown”.

The response prompt ”Below is a Python script with a self-contained function that solves the problem and passes corresponding tests:” translates to ”فيما يلي برنامج يحتوي على دالة بايثون مستقلة تحل المشكلة وتجتاز الاختبارات التالية:".

## A.3 Unit tests count distribution

We report in Figure 8 the histograms for unit test counting of HumanEval-Ar and MBPP-Ar.

## A.4 Cross-Task Correlation Analysis

To understand the possible correlation between the performance of an LLM on Arabic NLP, Arabic code, and English code benchmarks, we compute Pearson correlation coefficients between average evaluation scores across three tasks: Arabic NLP from the Open Arabic LLM Leaderboard[9] (OALL) (El Filali et al., 2025), English code generation from EvalPlus, and Arabic code generation from EvalPlus-Ar. Analysis includes only models evaluated on both code benchmarks and OALL (Table 6).

| | Arabic NLP | English Code | Arabic Code |
|---|---|---|---|
| **Arabic NLP** | 1.00 | 0.45 | 0.42 |
| **English Code** | 0.45 | 1.00 | 0.97 |
| **Arabic Code** | 0.42 | 0.97 | 1.00 |

Table 4: Pearson correlation between model scores across Arabic NLP, English code, and Arabic code tasks for base models.

**Base models:** English and Arabic code generation scores are tightly coupled ($r = 0.97$), indicating that code capabilities generalize well across languages when prompts are translated (Table 4, Figure 9). Arabic NLP shows moderate positive correlations with both English code ($r = 0.45$) and Arabic code ($r = 0.42$). Qwen models exhibit distinct behavior, achieving the best programming capabilities while dominating the upper-right quadrant with simultaneously high programming and Arabic-NLP scores (Figure 10).

| | Arabic NLP | English Code | Arabic Code |
|---|---|---|---|
| **Arabic NLP** | 1.00 | 0.10 | 0.24 |
| **English Code** | 0.10 | 1.00 | 0.97 |
| **Arabic Code** | 0.24 | 0.97 | 1.00 |

Table 5: Pearson correlation between model scores across Arabic NLP, English code, and Arabic code tasks for instruct models.

**Instruct models:** The tight coupling between English and Arabic code generation persists ($r = 0.97$), confirming that supervised fine-tuning preserves the underlying programming competence measured by both tracks (Table 5). However, the association between Arabic-NLP and code scores weakens considerably: Arabic NLP correlates only marginally with English code ($r = 0.10$) and modestly with Arabic code ($r = 0.24$). Figure 11 illustrates this decoupling through increased scatter across model families.

[9] https://huggingface.co/spaces/OALL/Open-Arabic-LLM-Leaderboard

These results suggest that instruct fine-tuning specializes models along specific objectives, reducing transferable overlap between programming skills and Arabic natural-language proficiency. The top-right quadrant features larger models (`Llama-3.3-70B-Instruct`, `Qwen-2.5-32B`, `Qwen-2.5-72B-Instruct`, `Gemma-3-27B-IT`), while Qwen models remain competitive on coding tasks even at smaller scales despite weaker Arabic NLP performance.

The near-perfect alignment between English and Arabic code scores contrasts with the moderate association between code and Arabic-NLP capabilities, reinforcing the need to evaluate these as complementary dimensions of LLM competence.

**Scores from Open-Arabic-LLM Leaderboard** We report in Tables 6, 7 (for base and instruct models, respectively) the average scores from Open-Arabic-LLM Leaderboard that are used to study the correlation between Arabic code generation, English code generation and Arabic NLP.

### A.5 Machine Translation

Figures 6 and 7 show rougeL-F1 distribution between the original and backtranslated prompts, before human check, for the HumanEval and MBPP benchmarks.

| Model | Size | Average |
|---|---|---|
| Qwen2.5 | 7B | 41.97 |
| | 14B | 54.26 |
| | 32B | 65.45 |
| | 72B | **69.37** |
| jais-adapted | 13B | 42.53 |
| | 70B | 51.94 |
| jais-family-8k | 30B | 53.63 |
| Fanar-1 | 9B | 62.83 |
| Llama-3.1 | 8B | 51.64 |
| AceGPT-v2 | 32B | 61.74 |
| | 70B | <u>67.20</u> |
| Qwen3-Base | 4B | 62.86 |
| | 8B | 66.22 |
| | 32B | 53.76 |
| gemma-3-pt | 27B | 63.20 |

Table 6: Base models performance on the Open-Arabic-LLM Leaderboard.

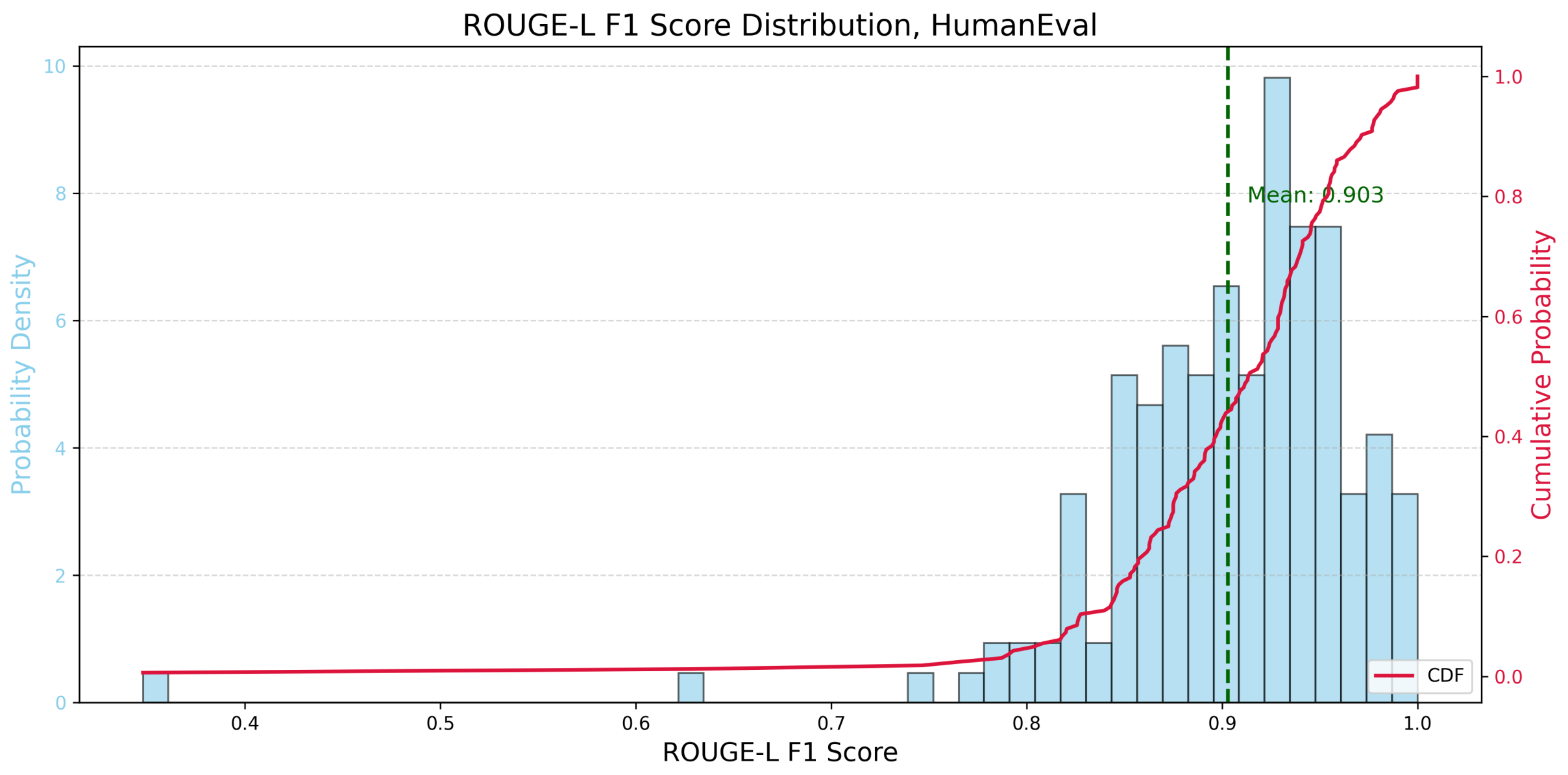

Figure 6: RougeL-f1 score distribution for round-trip translation of HumanEval input prompts, before human check.

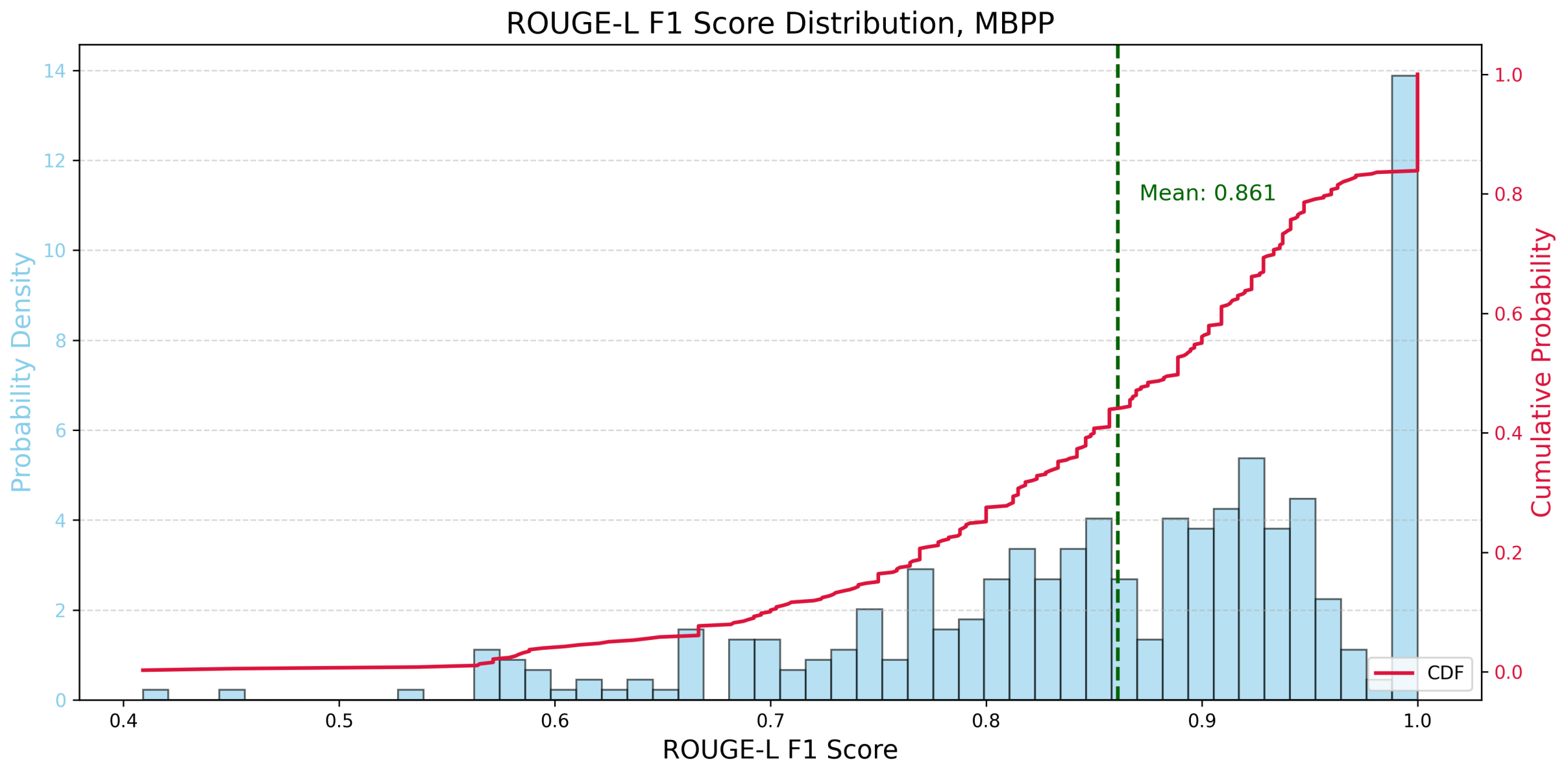

Figure 7: RougeL-f1 score distribution for round-trip translation of MBPP input prompts, before human check.

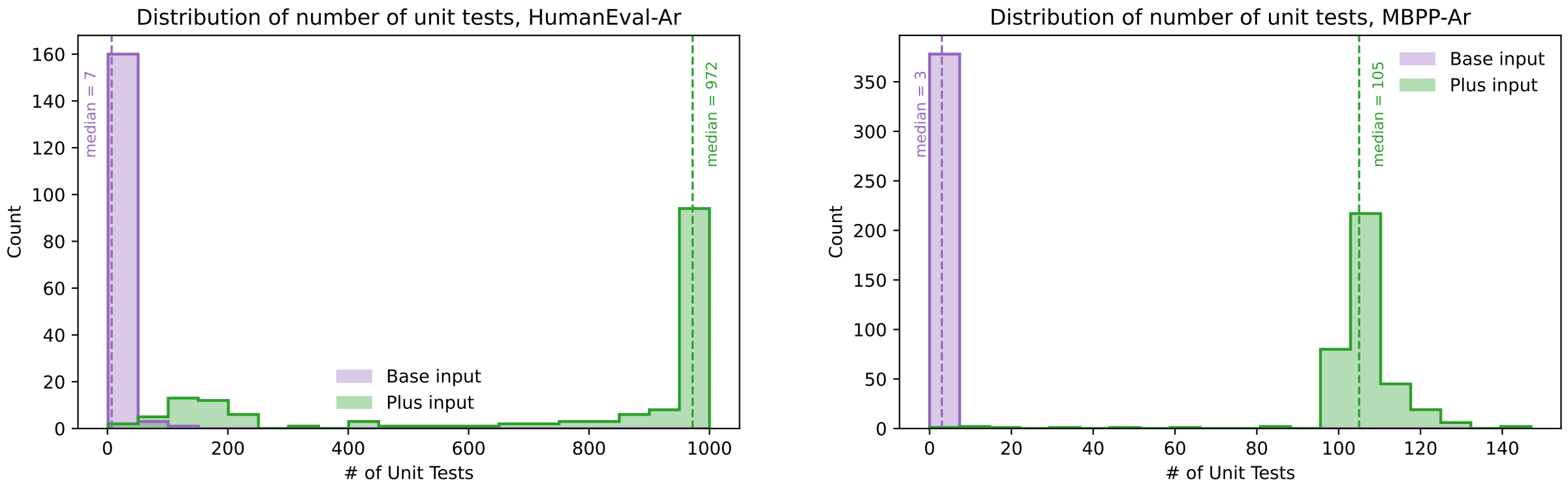

(a) HumanEval-Ar

(b) MBPP-Ar

Figure 8: Distribution of the number of unit tests for the benchmarks in the EvalPlus-Ar suite.

| Model | Size | Average |
|---|---|---|
| Qwen2.5-Instruct | 7B | 59.80 |
| | 14B | 63.18 |
| | 32B | 69.99 |
| | 72B | <u>72.39</u> |
| jais-adapted-chat | 13B | 58.08 |
| | 70B | 65.28 |
| SILMA-Instruct-v1.0 | 9B | 57.65 |
| Fanar-1-Instruct | 9B | 70.32 |
| Llama-3.1-Instruct | 8B | 55.41 |
| Llama-3.3-Instruct | 70B | **74.47** |
| AceGPT-v2-Chat | 8B | 62.35 |
| | 32B | 70.88 |
| | 70B | 70.07 |
| aya-expanse | 32B | 67.17 |
| c4ai-command-r-arabic-02-2025 | 7B | 67.07 |
| ALLaM-Instruct-preview | 7B | 65.25 |
| Yehia-preview | 7B | 65.68 |
| Qwen3 | 8B | 62.87 |
| | 14B | 45.34 |
| gemma-3-it | 27B | 71.4 |

Table 7: Instruct models performance on the Open-Arabic-LLM Leaderboard.

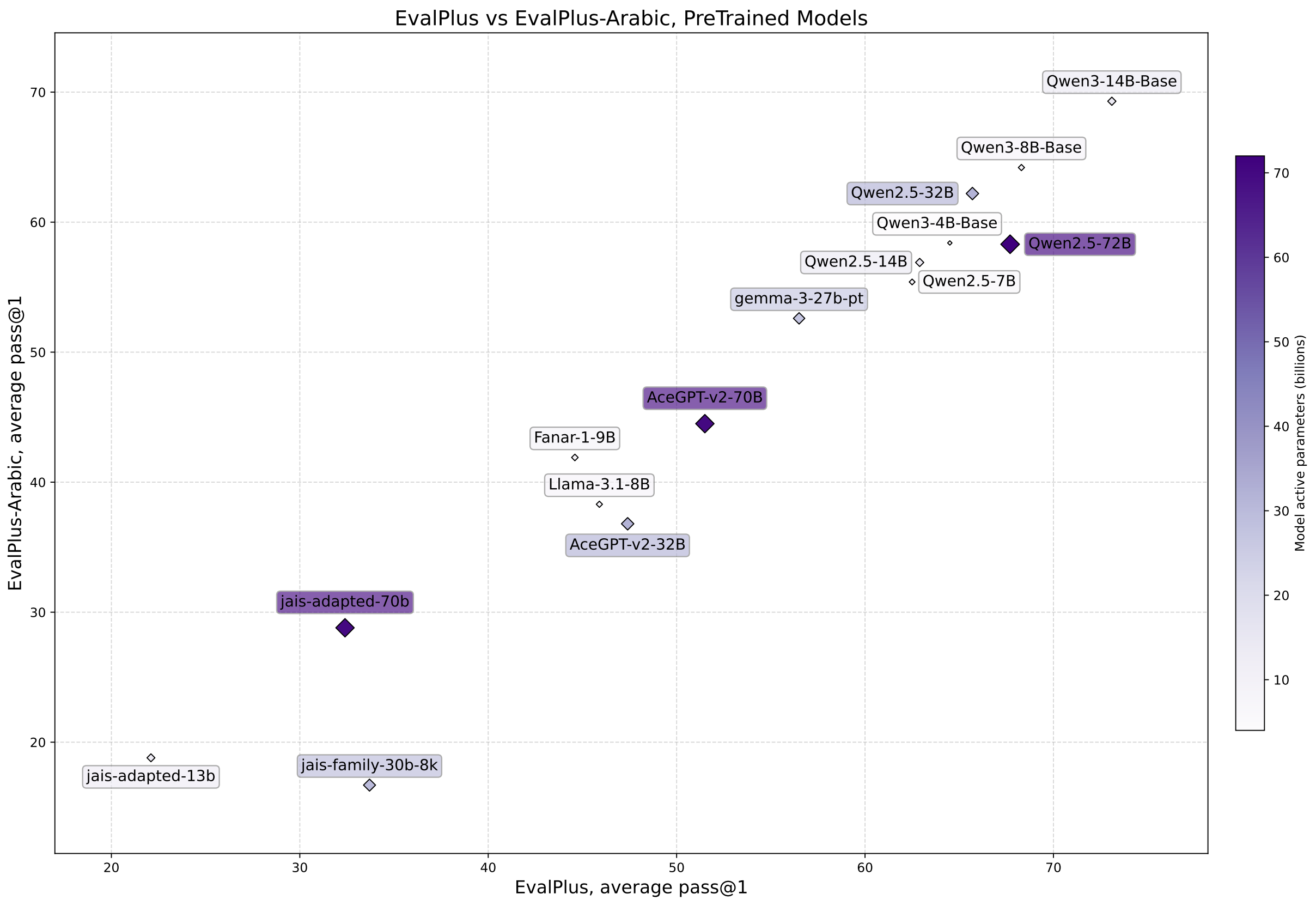

Figure 9: Correlation plot of EvalPlus and EvalPlus-Arabic suites for pre-trained models. Average pass@1 is reported as metric.

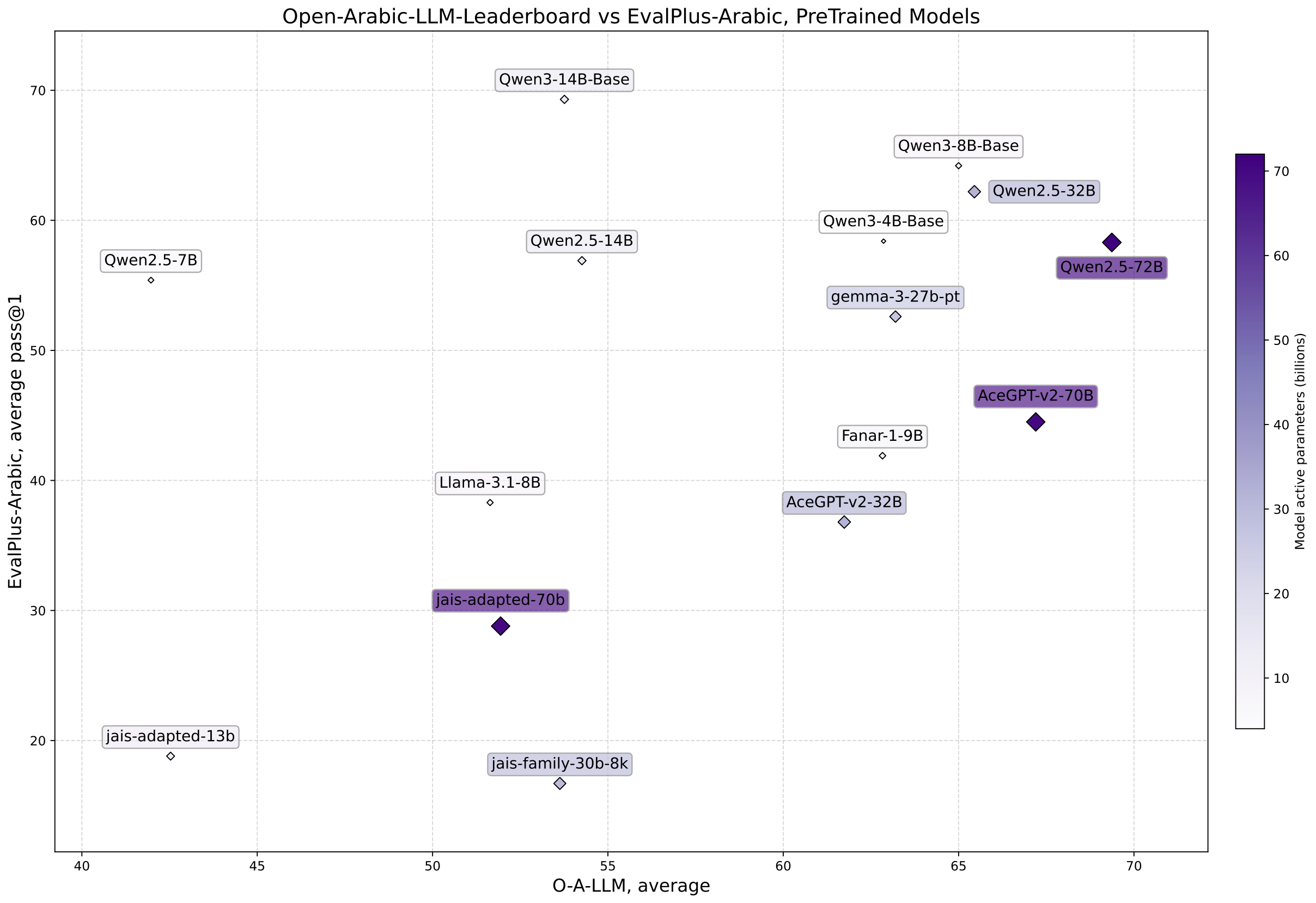

Figure 10: Correlation plot of OALL and EvalPlus-Arabic suites for pre-trained models. Average accuracy and average pass@1 are reported, respectively, as metrics.

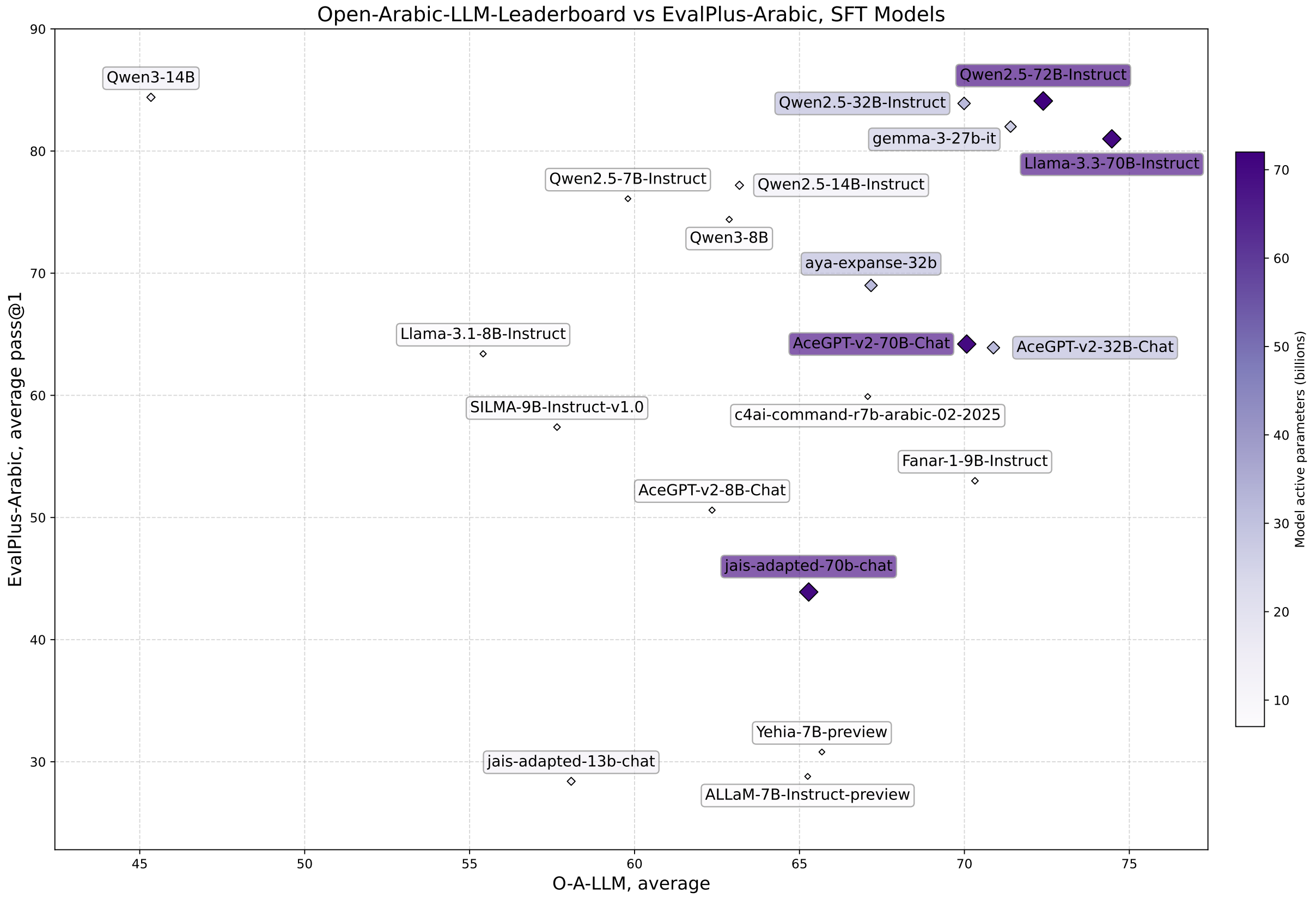

Figure 11: Correlation plot of OALL and EvalPlus-Arabic suites for instruct models. Average accuracy and average pass@1 are reported, respectively, as metrics.

# B Native Benchmark Prompts

## B.1 Prompt 1: Document QA Extraction

**Prompt 1**

You are given a document in Arabic extracted from an OCR-scanned source. Your task is to extract all **self-contained question–answer (QA) pairs** present in the text.

**Here is the document text:**

`{document}`

**Instructions:**

- Identify if there is a **global instruction or context** that applies to multiple questions (e.g., "Choose A or B", "Answer based on the paragraph above"). If such global context exists, **prepend it** to the relevant question so every question includes all necessary information to be understood independently.
- For **multiple choice questions**, include the full list of options directly in the question, clearly labeled (e.g., (A), (B), (C)), even if they appear across lines or pages.
- Match each question with its corresponding answer based on **labeling** (e.g., (1), (2), ) and positioning in the text.
- If no explicit answer is found nearby, check for an **answer table or list** at the end of the document and use it to assign the correct answer based on question number or label.
- For multiple choice questions, return only the **label of the correct option** (e.g., " ", "B", "3") — not the full text of the option.
- Ensure each question is **fully self-contained**, including any formatting or instructions needed to interpret it correctly.
- If a question refers to a figure, diagram, or drawing, include the full text — do **not skip** it automatically.

Your output should be a well-formed JSON object containing:

- A list of `qa_pairs`, where each entry includes:
  - `"question"`: Fully self-contained, with prepended global context if applicable.
  - `"answer"`: The corresponding answer, or empty string if none is found.

Return only the JSON output — do not include explanations, markdown, or extra text.
Handle possible OCR artifacts such as spelling variations, misplaced lines, or missing punctuation by interpreting the most likely intended meaning.

## B.2 Prompt 2: Question Classification and Metadata

**Prompt 2**

You are given a set of question–answer pairs from a school-level educational document. Your task is to classify each question by type, assign a difficulty score (1–10), identify the domain or subject, and determine if the question is visually dependent.

**Classify each question into one of these types:**

- `"MCQ"` — multiple choice question.
- `"Generative"` — open-ended explanation or description.
- `"Completion"` — fill-in-the-blank or short completion.
- `"Other"` — any other format not fitting above.

**Assign a difficulty score between 1 and 10**, where:

- 1 = very easy for high school students.
- 10 = very difficult for a high school graduate.

**Identify the subject or domain:**

- Chemistry / Biology / Physics / Math / History / Geography / Religion / Language / Other.

**Determine if the question is visually dependent:**

- `"is_visual"`: true if it refers to or asks for interpretation of figures, tables, plots, drawings, or instructs the student to draw or edit visuals.
- `"is_visual"`: false if the question is fully self-contained in text and does not require visual aids.

**Return a JSON object** with the same structure as input, but with added fields:

- `"type"`: "MCQ"/"Generative"/"Completion"/"Other".
- `"difficulty"`: integer 1–10.
- `"domain"`: e.g., "Chemistry".
- `"is_visual"`: boolean.

Do **NOT** include any extra text outside the JSON.

**Input:**

{input_data}

### B.3 Prompt 3: Final MCQ Formatting

**Prompt 3**

You are given a set of question–answer pairs in Arabic, extracted from Arabic OCR'd educational documents. Your task is to refine and enhance these pairs to be used in a high-quality dataset.

**Instructions:**

1. **Clean the format:**
   - Remove any explicit "Question:" or "Answer:" labels from both questions and answers.
   - If the pair is already an MCQ and appears clean (with clearly labeled options and a correct answer), leave both the question and answer unchanged.
2. **For non-MCQ pairs only:**
   - Generate a new **MCQ version** of the question based on the original content.
   - Include 4 options labeled as: "(أ)", "(ب)", "(ج)", "(د)".
   - One of the options must be the **correct answer**; assign it randomly among the four choices.
   - The remaining three options should be **plausible distractors**, related to the topic and context of the question.
   - Include both the correct choice label and the actual value in the `"answer"` field.
3. **Output structure:**
   - Return a list of JSON objects.
   - Each object should contain:
     - `"original_question"`: cleaned original question text (without labels).
     - `"original_answer"`: cleaned original answer text.
     - `"type"`: stays the same as the original type in input data.
     - `"refined_question"`: refined or generated MCQ question string, including all four options.
     - `"refined_answer"`: correct answer label and value.
     - `"refined"`: boolean True if changes were made, False if no refinement was needed.
     - `"difficulty"`: integer score from 1 (very easy) to 10 (very hard).
     - `"domain"`: subject or field (e.g., "History", "Math", etc.).
     - `"is_visual"`: boolean indicating if visual interpretation is needed.
4. Do **NOT** include any extra text outside the JSON output.

**Input:**

{input_data}

### B.4 Sample Questions

As shown in Figure 12, our benchmark includes both native and synthetic questions spanning various scientific domains such as biology, chemistry, mathematics, physics, and geography. This visual demonstrates not only the question formatting but also the attention to content diversity and difficulty calibration within our dataset. Additional details on the construction and classification of these questions are provided in Sections 3.1.2 and 3.1.3.

## C Question Type distribution across domains in Synthetic benchmark

Figures 13, 14, 15, 16 and 17 represent the domain-wise distribution of question types across the synthetic benchmark.

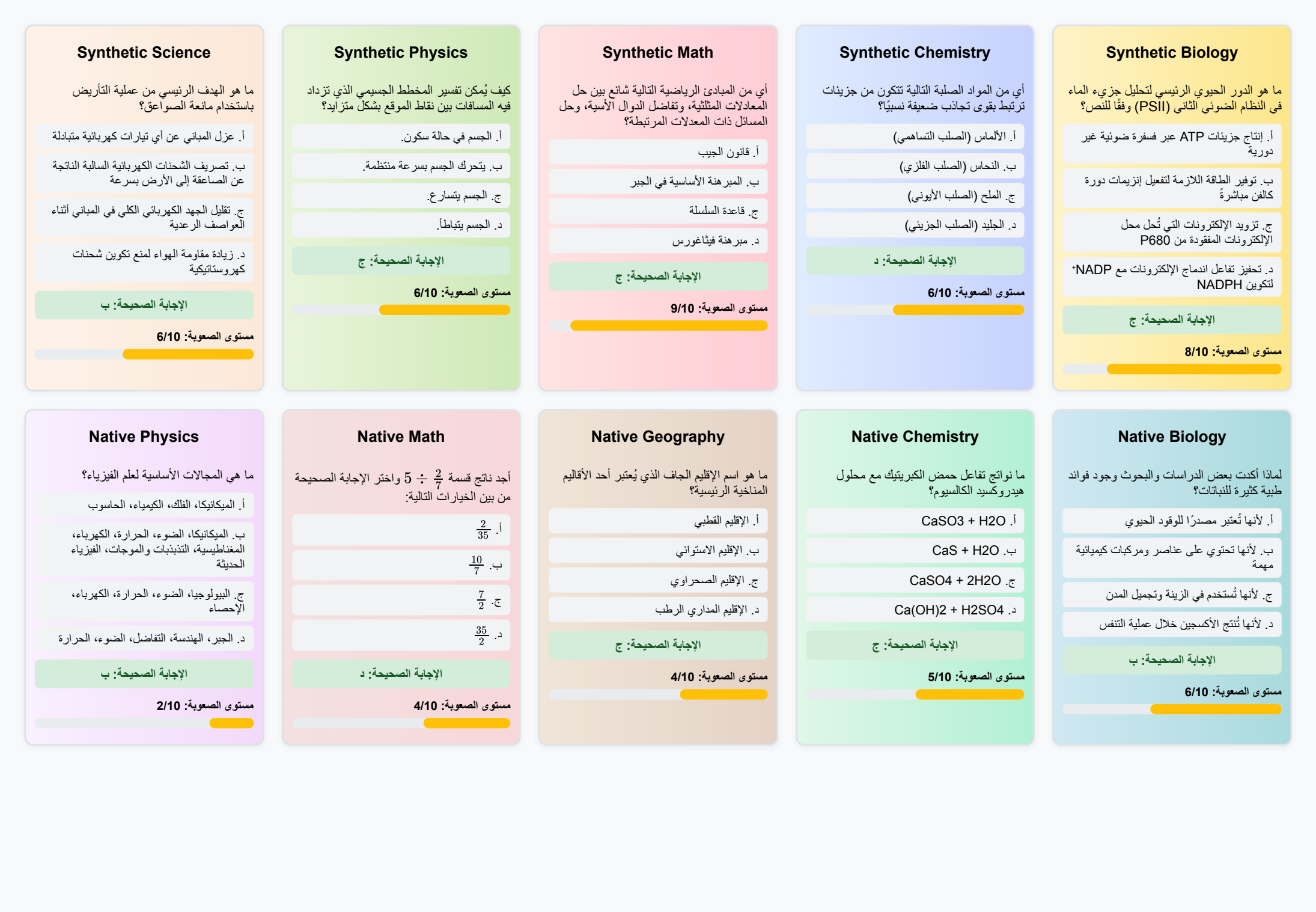

Figure 12: Examples of native and synthetic multiple-choice questions from the Arabic benchmark.

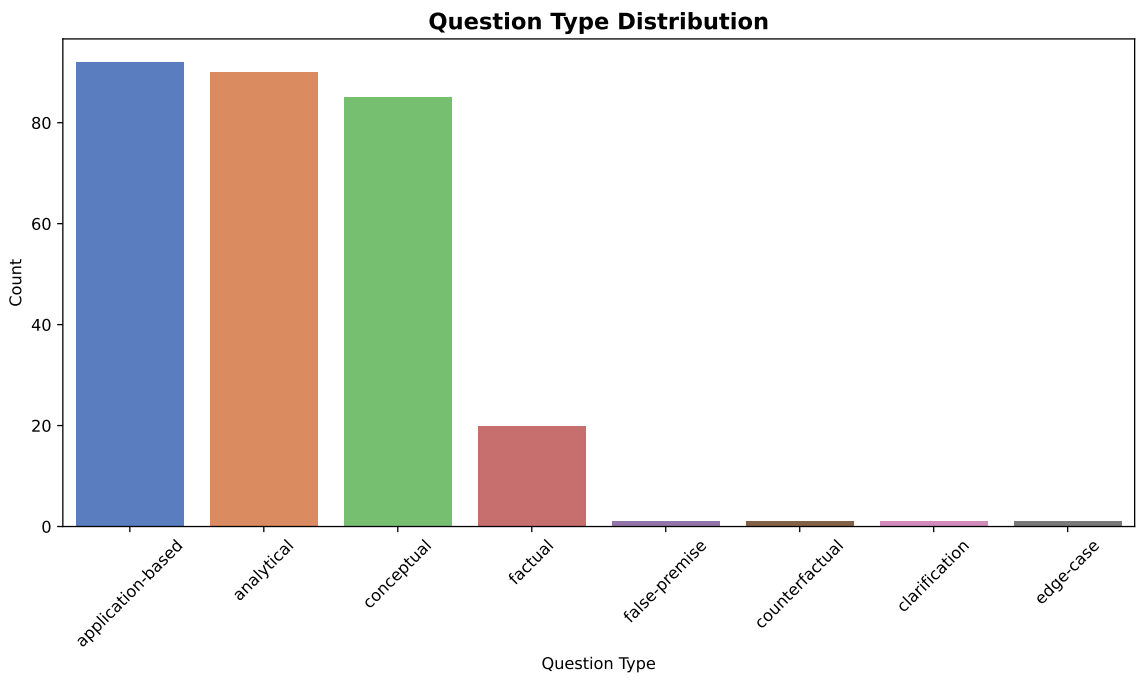

Figure 13: Biology question type distribution.

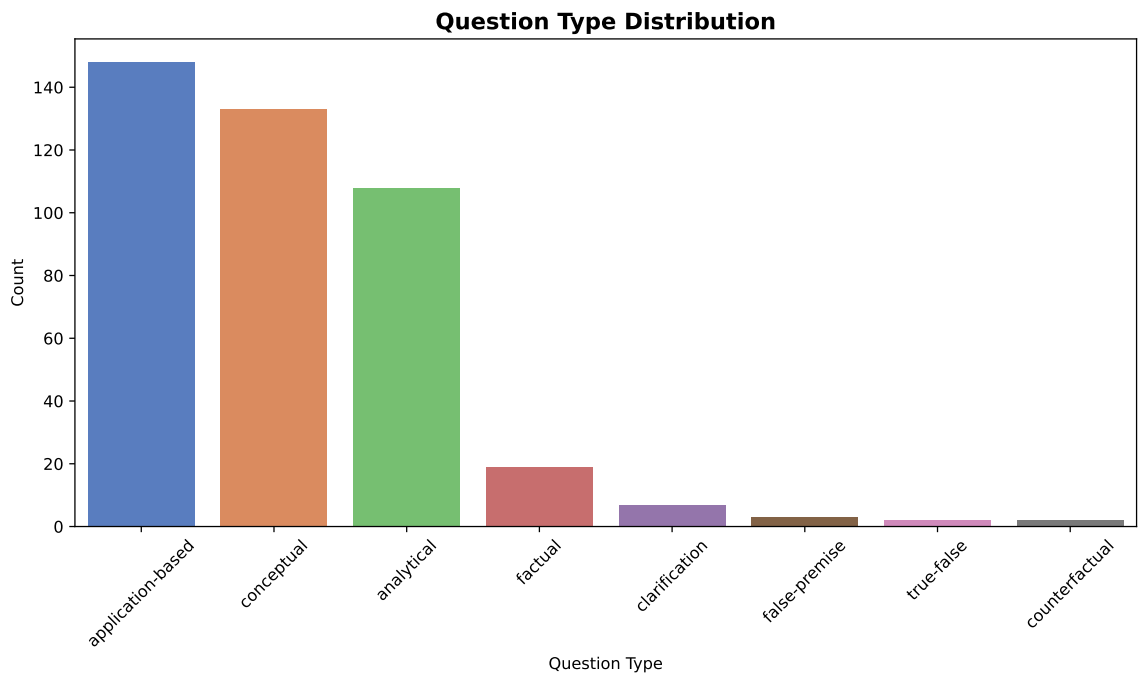

Figure 14: Chemistry question type distribution.

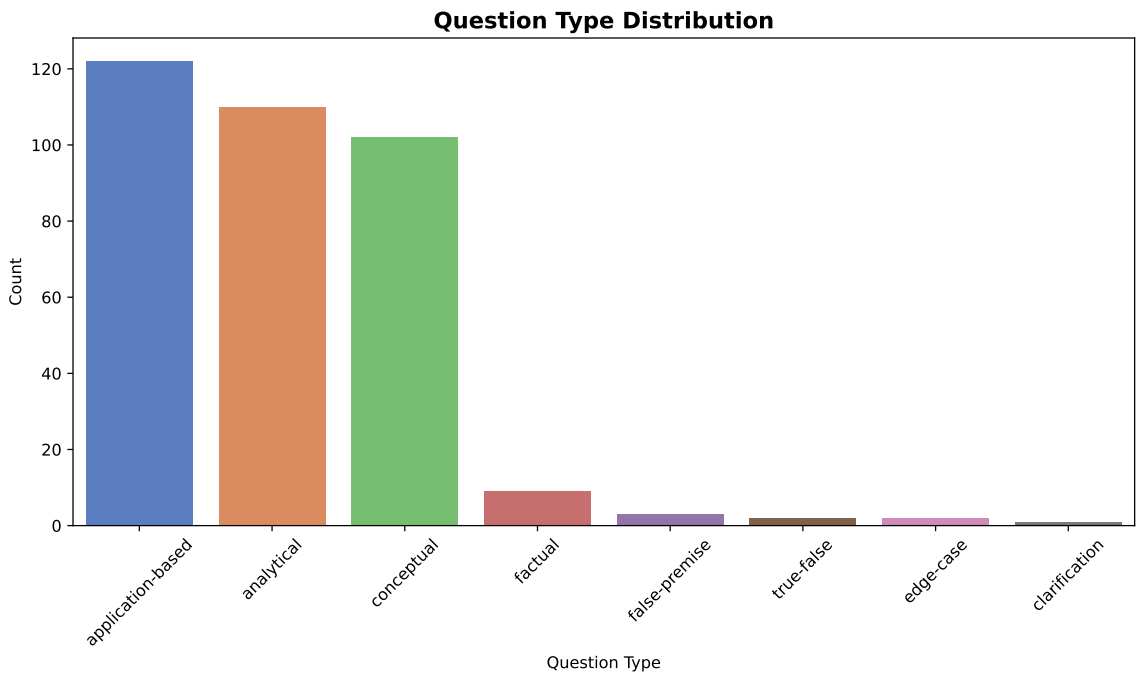

Figure 15: Math question type distribution.

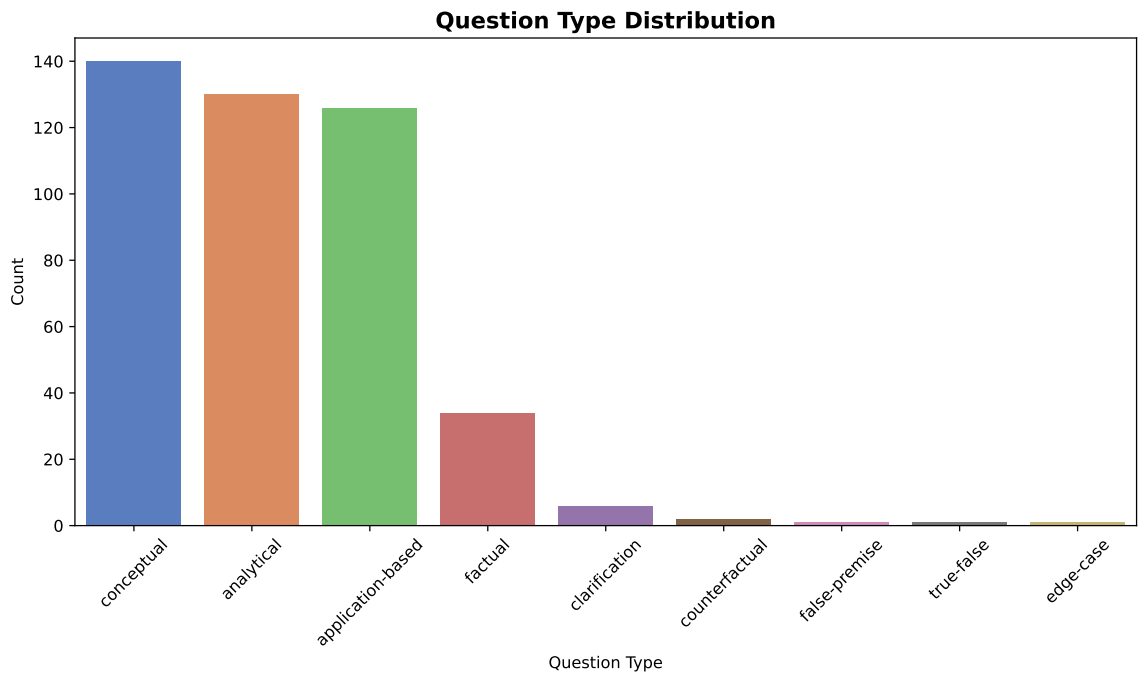

Figure 16: General Science question type distribution.

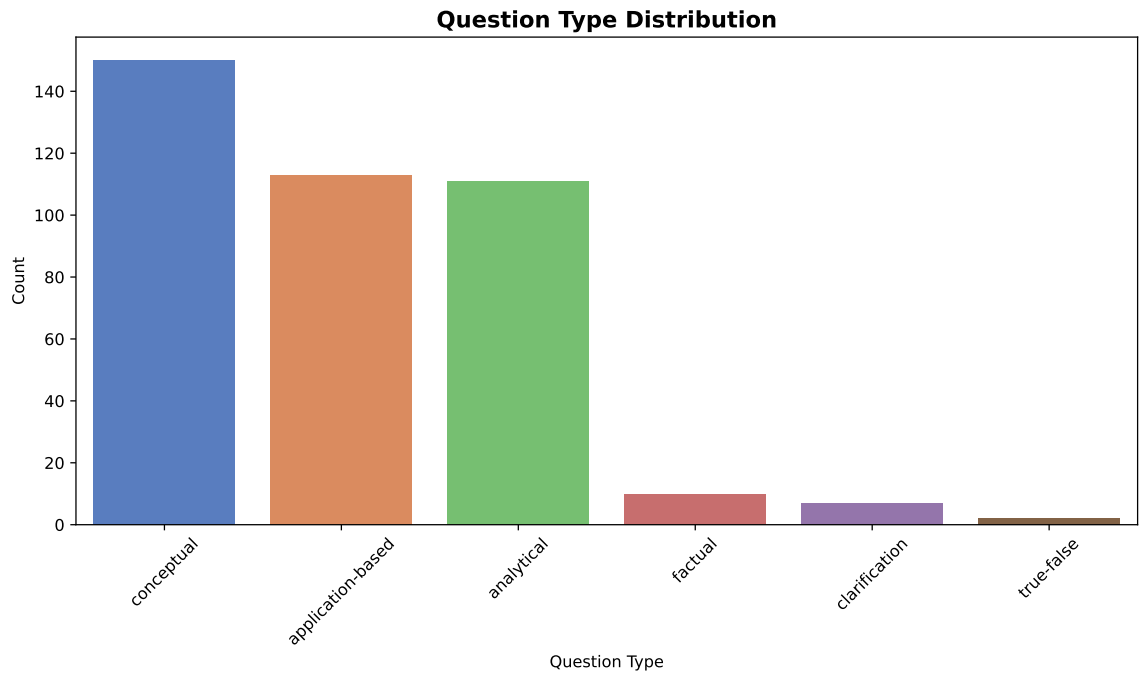

Figure 17: Physics question type distribution.

| Model | Size | Native MCQ | Native Completion | Synthetic MCQ | Synthetic Completion |
|---|---|---|---|---|---|
| Qwen2.5-Instruct | 7B | 62.65 | 51.32 | 79.50 | 44.94 |
| | 14B | 83.23 | 58.15 | 77.24 | 53.46 |
| | 32B | 89.36 | 63.12 | <u>86.35</u> | 58.10 |
| | 72B | **93.06** | 55.02 | **92.22** | **59.86** |
| jais-adapted-chat | 13B | 75.02 | 46.35 | 57.29 | 38.18 |
| | 70B | 73.29 | 50.28 | 70.41 | 44.52 |
| jais-chat-v3 | 30B | 78.95 | 56.88 | 62.98 | 40.92 |
| SILMA-Instruct-v1.0 | 9B | 86.7 | 59.88 | 77.92 | 52.03 |
| Fanar-1-Instruct | 9B | 89.24 | **67.39** | 82.81 | <u>59.28</u> |
| Llama-3.1-Instruct | 8B | 76.64 | 45.54 | 49.92 | 36.39 |
| Llama-3.3-Instruct | 70B | <u>92.60</u> | 61.61 | 86.18 | 55.18 |
| AceGPT-v2-Chat | 8B | 71.21 | 57.69 | 70.66 | 45.68 |
| | 32B | 90.17 | <u>65.89</u> | 82.50 | 47.98 |
| | 70B | 86.93 | 59.88 | 82.56 | 57.39 |
| aya-expanse | 8B | 80.34 | 56.06 | 61.91 | 41.86 |
| | 32B | 79.76 | 58.38 | 74.39 | 48.68 |
| c4ai-command-r-arabic-02-2025 | 7B | 79.19 | 52.48 | 67.51 | 41.87 |
| ALLaM-Instruct-preview | 7B | 81.15 | 61.38 | 71.01 | 53.05 |
| Yehia-preview | 7B | 82.08 | 62.77 | 70.63 | 49.74 |
| Qwen3 | 4B | 43.01 | 43.24 | 31.92 | 44.96 |
| | 8B | 20.23 | 47.63 | 30.76 | 47.34 |
| | 14B | 39.54 | 50.98 | 28.24 | 47.62 |
| | 32B | 29.02 | 53.87 | 35.80 | 52.80 |
| | 30B-A3B | 17.57 | 53.53 | 25.63 | 48.50 |
| | 235B-A22B | 65.78 | 55.49 | 29.85 | 56.47 |
| gemma-3-it | 4B | 49.82 | 49.13 | 31.96 | 44.20 |
| | 12B | 90.86 | 64.04 | 82.41 | 55.63 |
| | 27B | 91.56 | 63.69 | 80.42 | 58.37 |

Table 8: MCQ vs. Completion average for instruct models. **Bold** indicates the highest score in each column; <u>Underline</u> indicates the second best.

| Model | Size | MCQ Biology | MCQ Chemistry | MCQ General Science | MCQ Math | MCQ Physics | Completion Biology | Completion Chemistry | Completion General Science | Completion Math | Completion Physics |
|---|---|---|---|---|---|---|---|---|---|---|---|
| Qwen2.5-Instruct | 7B | 84.93 | 72.22 | 85.14 | 77.39 | 77.81 | 39.34 | 41.3 | 50.54 | 51.27 | 42.25 |
| | 14B | 84.19 | 75.6 | 88.92 | 62.1 | 75.4 | 54.78 | 52.42 | 55.95 | 52.55 | 51.6 |
| | 32B | 90.07 | <u>85.75</u> | 90.27 | <u>79.3</u> | <u>86.36</u> | 56.25 | 58.94 | 58.65 | 58.92 | 57.75 |
| | 72B | **96.69** | **89.37** | **96.49** | **86.31** | **92.25** | 58.82 | 59.18 | 61.62 | <u>60.83</u> | **58.82** |
| Jais-adapted | 13B | 72.43 | 52.42 | 70.81 | 35.99 | 54.81 | 41.54 | 37.92 | 44.05 | 31.53 | 35.83 |
| | 70B | 74.26 | 59.42 | 75.95 | 47.8 | 57.49 | 45.96 | 43.48 | 51.89 | 43.31 | 37.97 |
| Jais-chat-v3 | 30B | 79.78 | 67.39 | 81.62 | 51.59 | 71.66 | 43.75 | 35.99 | 52.7 | 35.03 | 37.16 |
| SILMA-Instruct-v1.0 | 9B | 84.19 | 78.99 | 85.68 | 67.2 | 73.53 | 52.21 | 50 | 52.16 | 54.46 | 51.34 |
| Fanar-1-Instruct | 9B | 90.07 | 80.43 | 89.73 | 72.29 | 81.55 | **64.71** | 57 | 62.43 | 54.78 | 57.49 |
| Llama-3-Instruct | 8B | 56.99 | 48.07 | 62.7 | 35.03 | 46.79 | 39.34 | 30.92 | 41.89 | 36.62 | 33.16 |
| | 70B | <u>93.38</u> | 83.33 | <u>94.86</u> | 73.25 | 86.1 | 55.15 | 53.86 | 55.14 | **61.46** | 50.27 |
| AceGPT-v2-Chat | 8B | 80.88 | 67.15 | 82.97 | 57.6 | 64.71 | 48.53 | 42.03 | 51.35 | 42.36 | 44.12 |
| | 32B | 88.6 | 81.16 | 90 | 71.97 | 80.75 | 46.32 | 49.03 | 52.16 | 49.04 | 43.32 |
| | 70B | 90.07 | 81.4 | 91.08 | 69.75 | 80.48 | 58.46 | 53.62 | **63.24** | 55.73 | 55.88 |
| aya-expanse | 8B | 73.53 | 56.04 | 75.68 | 45.22 | 59.09 | 46.69 | 36.71 | 48.11 | 39.81 | 37.97 |
| | 32B | 87.87 | 74.15 | 85.68 | 51.27 | 72.99 | 52.57 | 46.14 | 54.05 | 46.5 | 44.12 |
| c4ai-command-r7b-arabic-02-2025 | 7B | 76.47 | 64.49 | 81.08 | 51.59 | 63.9 | 47.79 | 34.06 | 47.3 | 41.72 | 38.5 |
| ALLaM-Instruct-preview | 7B | 78.31 | 70.29 | 86.22 | 52.87 | 67.38 | <u>58.82</u> | 50 | <u>62.97</u> | 42.68 | 50.8 |
| Yehia-preview | 7B | 77.57 | 69.08 | 85.95 | 53.18 | 67.38 | 52.94 | 45.65 | 58.38 | 41.72 | 50 |
| Qwen3 | 4B | 30.88 | 35.27 | 34.59 | 26.75 | 32.09 | 41.54 | 43.72 | 47.03 | 44.9 | 47.59 |
| | 8B | 28.31 | 32.85 | 32.7 | 28.66 | 31.28 | 45.96 | 49.28 | 43.78 | 49.04 | 48.66 |
| | 14B | 25.74 | 31.16 | 29.19 | 25.16 | 29.95 | 44.12 | 48.79 | 47.57 | 48.41 | 49.2 |
| | 32B | 35.66 | 39.13 | 37.57 | 28.66 | 37.97 | 51.84 | 52.9 | 52.97 | 52.55 | 53.74 |
| | 30B-A3B | 22.06 | 28.02 | 25.14 | 24.84 | 28.07 | 49.26 | 51.21 | 45.41 | 49.04 | 47.59 |
| | 235B-A22B | 32.35 | 28.5 | 34.59 | 25.48 | 28.34 | 55.88 | **59.9** | 55.14 | 52.87 | <u>58.56</u> |
| gemma-3 | 4B | 29.04 | 32.13 | 33.24 | 34.39 | 31.02 | 43.75 | 41.3 | 45.95 | 47.77 | 42.25 |
| | 12B | 90.81 | 80.43 | 89.19 | 71.66 | 79.95 | 55.51 | 57.97 | 56.76 | 56.05 | 51.87 |
| | 27B | 87.13 | 80.92 | 83.51 | 73.25 | 77.27 | 58.09 | <u>59.42</u> | 58.65 | 59.55 | 56.15 |

Table 9: Instruct models performance on the synthetic benchmark (values in percentages). **Bold** indicates the best score in each column; <u>underline</u> indicates the second best.

| | | English | | | | Arabic | | | | Average | |
|---|---|---|---|---|---|---|---|---|---|---|---|
| Model | Size | HumanEval | HumanEval+ | MBPP | MBPP+ | HumanEval | HumanEval+ | MBPP | MBPP+ | English | Arabic |
| Qwen2.5-Instruct | 7B | 82.3 | 74.4 | 79.1 | 68.5 | 73.2 | 66.5 | 78.0 | 67.2 | 76.1 | 71.2 |
| | 14B | 82.3 | 75.0 | 82.0 | 69.3 | 72.6 | 65.2 | 78.6 | 65.3 | 77.2 | 70.4 |
| | 32B | 89.0 | 82.3 | 88.9 | 75.4 | 82.3 | 75.0 | 84.9 | 71.4 | 83.9 | 78.4 |
| | 72B | 87.8 | 81.7 | **90.2** | **76.5** | 83.5 | 76.2 | **87.0** | 72.5 | 84.1 | 79.8 |
| jais-adapted-chat | 13B | 21.3 | 18.3 | 40.5 | 33.3 | 11.0 | 10.4 | 29.6 | 24.3 | 28.4 | 18.8 |
| | 70B | 39.0 | 34.1 | 55.3 | 47.1 | 17.7 | 15.2 | 41.3 | 34.7 | 43.9 | 27.2 |
| jais-chat-v3 | 30B | 26.2 | 23.2 | 36.2 | 30.4 | 22.0 | 18.9 | 28.3 | 24.3 | 29.0 | 23.4 |
| SILMA-Instruct-v1.0 | 9B | 53.7 | 48.8 | 69.3 | 57.9 | 46.3 | 38.4 | 62.2 | 53.4 | 57.4 | 50.1 |
| Fanar-1-Instruct-1 | 9B | 63.4 | 54.3 | 50.0 | 44.4 | 54.3 | 45.7 | 47.6 | 40.5 | 53.0 | 47.0 |
| Llama-3.1-Instruct | 8B | 68.9 | 62.2 | 67.5 | 54.8 | 49.4 | 43.3 | 56.1 | 48.4 | 63.4 | 49.3 |
| Llama-3.3-Instruct | 70B | 84.1 | 78.7 | 87.8 | 73.5 | 81.7 | 73.8 | 86.5 | 70.9 | 81.0 | 78.2 |
| AceGPT-v2-Chat | 8B | 47.0 | 41.5 | 62.4 | 51.6 | 37.2 | 30.5 | 54.2 | 45.8 | 50.6 | 41.9 |
| | 32B | 69.5 | 62.2 | 66.9 | 57.1 | 56.7 | 49.4 | 62.7 | 51.3 | 63.9 | 55.0 |
| | 70B | 64.6 | 57.3 | 73.3 | 61.6 | 55.5 | 48.8 | 72.8 | 60.6 | 64.2 | 59.4 |
| aya-expanse | 8B | 42.7 | 37.8 | 65.1 | 56.9 | 37.2 | 31.1 | 59.3 | 50.8 | 50.6 | 44.6 |
| | 32B | 70.7 | 64.0 | 75.7 | 65.6 | 5.5 | 49.4 | 65.6 | 56.3 | 69.0 | 56.7 |
| c4ai-command-r-arabic-02-2025 | 7B | 59.8 | 52.4 | 69.0 | 58.5 | 51.8 | 45.7 | 63.8 | 54.8 | 59.9 | 54.0 |
| ALLaM-Instruct-preview | 7B | 24.4 | 21.3 | 37.3 | 32.3 | 28.0 | 23.8 | 39.4 | 33.6 | 28.8 | 31.2 |
| Yehia-preview | 7B | 26.2 | 22.6 | 40.5 | 33.9 | 26.8 | 22.6 | 40.2 | 32.8 | 30.8 | 30.6 |
| Qwen3 | 4B | 82.9 | 76.2 | 70.1 | 60.8 | 74.4 | 65.2 | 70.1 | 58.5 | 72.5 | 67.1 |
| | 8B | 84.8 | 79.3 | 71.4 | 61.9 | 79.9 | 74.4 | 53.7 | 46.0 | 74.4 | 63.5 |
| | 14B | 88.4 | 86.0 | 87.3 | 75.7 | 82.3 | 76.8 | 61.4 | 52.6 | 84.4 | 68.3 |
| | 32B | 87.8 | 81.1 | **90.2** | **76.5** | 83.5 | 76.8 | 86.8 | **72.8** | 83.9 | **80.0** |
| | 30B-A3B | **94.5** | **89.0** | 86.0 | 73.5 | 83.5 | 78.0 | 54.0 | 45.8 | **85.8** | 65.3 |
| | 235B-A22B | 90.2 | 81.7 | 83.1 | 70.1 | 85.4 | **81.7** | 81.5 | 69.6 | 81.3 | 79.6 |
| gemma-3-it | 4B | 66.5 | 61.6 | 78.3 | 68.0 | 61.0 | 54.9 | 65.3 | 55.8 | 68.6 | 59.3 |
| | 12B | 84.8 | 76.2 | 85.4 | 71.7 | 79.9 | 73.2 | 83.6 | 70.4 | 79.5 | 76.8 |
| | 27B | 87.2 | 78.0 | 88.4 | 74.3 | **86.0** | 69.3 | 84.7 | 69.6 | 82.0 | 77.4 |

Table 10: Instruct models performance on the EvalPlus suite. **Bold** indicates the highest score in each column; Underline indicates the second highest.

| Model | Size | MCQ | | Completion | |
|---|---|---|---|---|---|
| | | Score | 25% | Score | 25% |
| Qwen2.5 | 7B | 86.13 | 77.57 | 48.43 | 41.27 |
| | 14B | 89.82 | 80.69 | 55.37 | 46.70 |
| | 32B | 93.41 | 83.70 | 56.18 | 47.51 |
| | 72B | **94.45** | 85.43 | 62.31 | 51.32 |
| jais-adapted | 13B | 43.81 | 38.49 | 57.91 | 47.97 |
| | 70B | 65.20 | 56.07 | 60.58 | 50.17 |
| jais-family-8k | 30B | 74.10 | 61.04 | 58.15 | 48.55 |
| Fanar-1 | 9B | 88.32 | 76.53 | 60.11 | 50.17 |
| Llama-3.1 | 8B | 73.52 | 65.78 | 45.78 | 37.57 |
| | 70B | 62.89 | 55.95 | 61.50 | 51.45 |
| AceGPT-v2 | 8B | 74.57 | 62.08 | 53.64 | 45.20 |
| | 32B | 81.27 | 70.64 | 55.95 | 47.16 |
| | 70B | 90.17 | 80.11 | 60.69 | 51.67 |
| Qwen3-Base | 4B | 87.05 | 77.69 | 48.32 | 42.66 |
| | 8B | 90.98 | 80.12 | 46.82 | 40.00 |
| | 14B | 87.98 | 77.46 | 50.98 | 43.47 |
| | 30B | 94.10 | **86.36** | 60.12 | 50.29 |
| gemma-3-pt | 4B | 81.15 | 66.12 | 52.02 | 43.93 |
| | 12B | 89.47 | 77.22 | 61.50 | 51.32 |
| | 27B | 94.10 | 83.93 | **67.63** | **56.18** |

Table 11: Native benchmark results for base models. **Bold** indicates the highest score in each column; Underline indicates the second best.

| Model | Size | MCQ | | Completion | |
|---|---|---|---|---|---|
| | | Score | 25% | Score | 25% |
| Qwen2.5-Instruct | 7B | 62.65 | 60.46 | 51.32 | 43.23 |
| | 14B | 83.23 | 72.71 | 58.15 | 49.82 |
| | 32B | 89.36 | <u>84.16</u> | 63.12 | 54.91 |
| | 72B | **93.06** | **93.06** | 55.02 | **64.97** |
| jais-adapted-chat | 13B | 75.02 | 68.32 | 46.35 | 39.19 |
| | 70B | 73.29 | 73.29 | 50.28 | 41.50 |
| jais-chat-v3 | 30B | 78.95 | 71.56 | 56.88 | 49.02 |
| SILMA-Instruct-v1.0 | 9B | 86.70 | 76.99 | 59.88 | 49.24 |
| Fanar-1-Instruct | 9B | 89.24 | 80.46 | **67.39** | <u>55.83</u> |
| Llama-3.1-Instruct | 8B | 76.64 | 69.47 | 45.54 | 37.34 |
| Llama-3.3-Instruct | 70B | <u>92.60</u> | 83.46 | 61.61 | 50.17 |
| AceGPT-v2-Chat | 8B | 71.21 | 67.86 | 57.69 | 48.44 |
| | 32B | 90.17 | 80.80 | 59.88 | 49.71 |
| | 70B | 86.93 | 80.00 | <u>65.89</u> | 55.37 |
| aya-expanse | 8B | 80.34 | 71.79 | 56.06 | 47.16 |
| | 32B | 79.76 | 72.02 | 58.38 | 49.82 |
| c4ai-command-r-arabic-02-2025 | 7B | 79.19 | 70.86 | 52.48 | 43.69 |
| ALLaM-Instruct-preview | 7B | 81.15 | 69.13 | 61.38 | 51.90 |
| Yehia-preview | 7B | 82.08 | 69.94 | 62.77 | 53.17 |
| Qwen3 | 4B | 43.01 | 40.81 | 43.24 | 38.03 |
| | 8B | 20.23 | 19.42 | 47.63 | 41.62 |
| | 14B | 39.54 | 35.03 | 50.98 | 43.12 |
| | 30B | 29.02 | 27.28 | 53.87 | 45.43 |
| | 30B-A3B | 17.57 | 16.99 | 53.53 | 46.94 |
| | 235B-A22B | 65.78 | 60.58 | 55.49 | 49.83 |
| gemma-3-it | 4B | 49.82 | 43.12 | 49.13 | 42.31 |
| | 12B | 90.86 | 78.72 | 64.04 | 54.91 |
| | 27B | 91.56 | 80.69 | 63.69 | 52.83 |

Table 12: Native benchmark results for instruct models. **Bold** indicates the highest score in each column; <u>Underline</u> indicates the second best.